\documentclass{article}
\usepackage{arxiv}

\usepackage[utf8]{inputenc} 
\usepackage[T1]{fontenc}    
\usepackage{hyperref}       
\usepackage{url}         
\usepackage{booktabs}      
\usepackage{amsfonts}       
\usepackage{nicefrac}       
\usepackage{microtype}      
\usepackage{graphicx}

\usepackage{tabularx}
\usepackage{adjustbox}
\usepackage{subcaption}
\usepackage{multicol}
\usepackage{lipsum}

\title{Linguistic Fingerprint in Transformer Models: \\ How Language Variation Influences Parameter Selection in Irony Detection}

\author{
Michele Mastromattei \\
	Campus Bio-Medico University of Rome, Italy \\
    University of Rome Tor Vergata, Italy \\
    \texttt{michele.mastromattei@\{unicampus, uniroma2\}.it}\\
	\And
	Fabio Massimo Zanzotto \\
    University of Rome Tor Vergata, Italy \\
}

\begin{document}
\maketitle

\begin{abstract}
This paper explores the correlation between linguistic diversity, sentiment analysis and transformer model architectures. We aim to investigate how different English variations impact transformer-based models for irony detection. To conduct our study, we used the EPIC corpus to extract five diverse English variation-specific datasets and applied the KEN pruning algorithm on five different architectures. Our results reveal several similarities between optimal subnetworks, which provide insights into the linguistic variations that share strong resemblances and those that exhibit greater dissimilarities. We discovered that optimal subnetworks across models share at least 60\% of their parameters, emphasizing the significance of parameter values in capturing and interpreting linguistic variations. This study highlights the inherent structural similarities between models trained on different variants of the same language and also the critical role of parameter values in capturing these nuances.
\end{abstract}


\section{Introduction} \label{sec: introduction}
Sentiment analysis datasets, particularly those annotated on crowdsourcing platforms, may contain biases due to the lack of information about the cultural backgrounds of the annotators. This can lead to machine learning models trained on this data amplifying these biases, affecting how people perceive and label sentiment. Although these models can capture general sentiment, they often fail to capture the nuances experienced by different groups.

This paper examines the impact of linguistic diversity on transformer models designed for irony detection. Using the EPIC corpus \cite{frenda-etal-2023-epic}, we created five subsets tailored to different variations of English. We trained different transformer models and used the KEN pruning algorithm \cite{mastromattei2024less} to extract the minimum subset of optimal parameters that maintain the original performance of the model. We conducted this experimental process across five transformer architectures, revealing a minimum parameter overlap of 60\% among resulting subnetworks. We then performed a comprehensive analysis to identify subnetworks with the highest and lowest similarity. Additionally, we used $KEN_{viz}$ for a visual examination of pattern similarities. Our results show that the linguistic variation is closely related to the individual values of each parameter within the models. This suggests that the diversity among linguistic variation is not just a structural aspect, but is deeply rooted in the specific values contained in the model. These insights can help create models that better capture the richness of linguistic variation and address bias effectively.
\newpage
\begin{figure*}[h!]
    \centering
    \includegraphics[width=\linewidth]{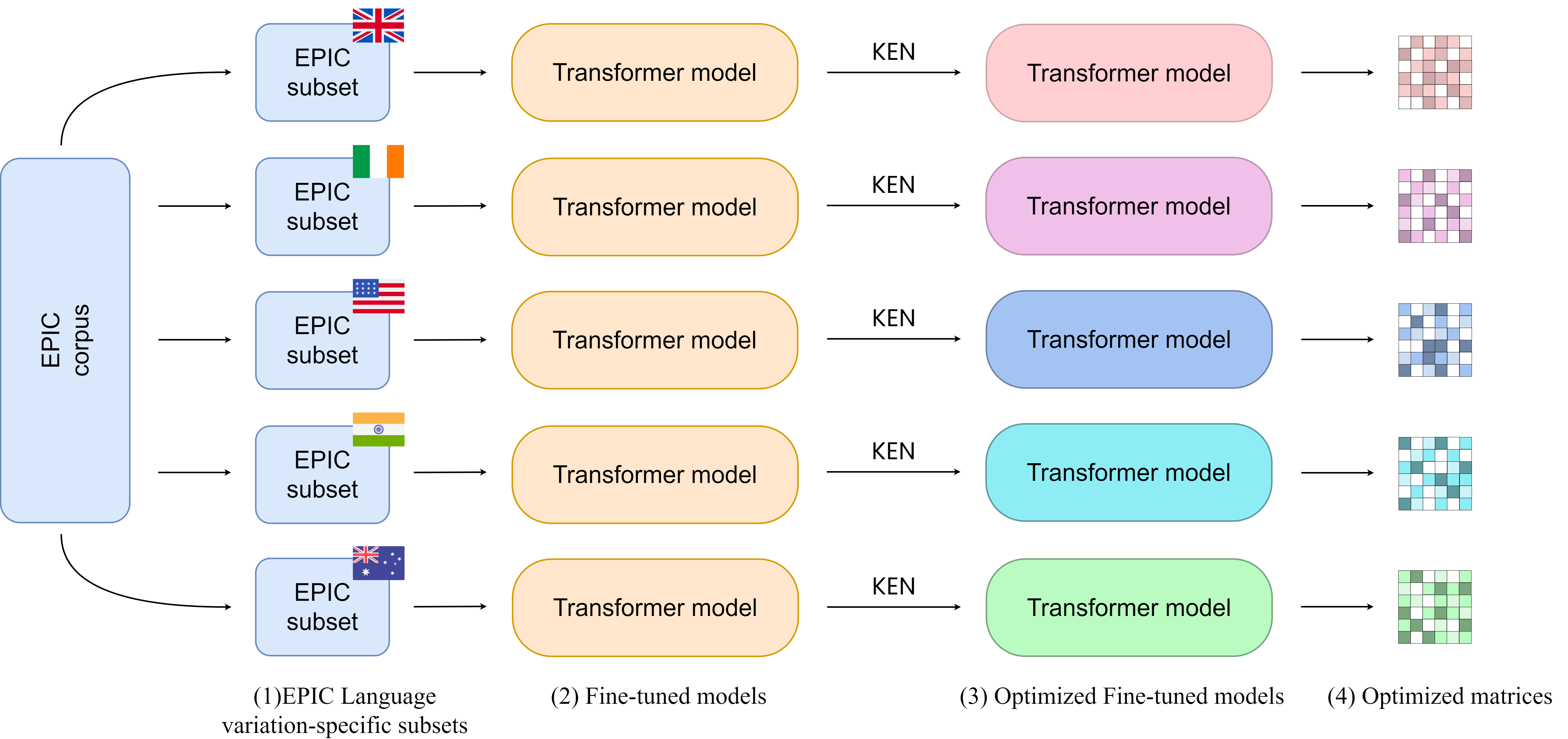}
    \caption{Workflow overview. Specific language variations are selected from the EPIC corpus (1). For each unique language subset, a dedicated transformer model is trained (2). This ensures that each model specializes in the intricacies of its assigned language variation. Finally, the KEN pruning algorithm is applied to optimize the trained models (3).   This involves efficient and lightweight architectures for each language variant (4).}
    \label{fig:workpath}
\end{figure*}

\section{Background and related work} \label{sec: Background}

Artificial intelligence (AI) models impact our daily lives in many ways. Some applications go beyond just processing data and strive to understand the intricate human elements and cultural nuances of our world. For instance, sentiment analysis requires a deeper understanding of implicit phrases and cultural differences to accurately interpret emotions \cite{tourimpampa2018perception, sun2022investigating}. This is why rigorous studies are essential before deploying data and models in real-world settings.
When creating data, it is crucial to incorporate different perspectives evaluation standards, such as "golden standards" \cite{basile2021toward}, incorporating criteria for evaluating annotators \cite{milkowski2021personal, abercrombie2023consistency, mieleszczenko2023capturing}, grouping them according to potential bias factors \cite{fell2021mining} or using text visualization techniques to analyze annotated datasets \cite{havens2022beyond}. On the model level, explainable AI (XAI) techniques \cite{samek2017explainable, samek2019towards, vilone2021notions} are being used to demystify complex models and ensure transparency. Many neural interpretability models rely on attention-based techniques \cite{bodria2020explainability}, utilizing auxiliary tasks \cite{de2019using}, or external knowledge integration \cite{zhao2021knowledge}.
Moreover, attention-based models exhibit a grasp of the syntactic structure of analyzed sentences \cite{manning2020emergent}. Consequently, the role of syntax in model interpretation is being extensively studied across various domains, including irony \cite{cignarella2020multilingual} and hate speech \cite{mastromattei2022syntax, mastromattei2022change}. This multifaceted exploration contributes to a richer understanding of the interplay between language, culture and model interpretability to achieve increasingly inclusive AI models.
\section{Methods and Data} \label{sec: Methods}

\begin{figure*}[h!]
    \centering
    \includegraphics[width=\linewidth]{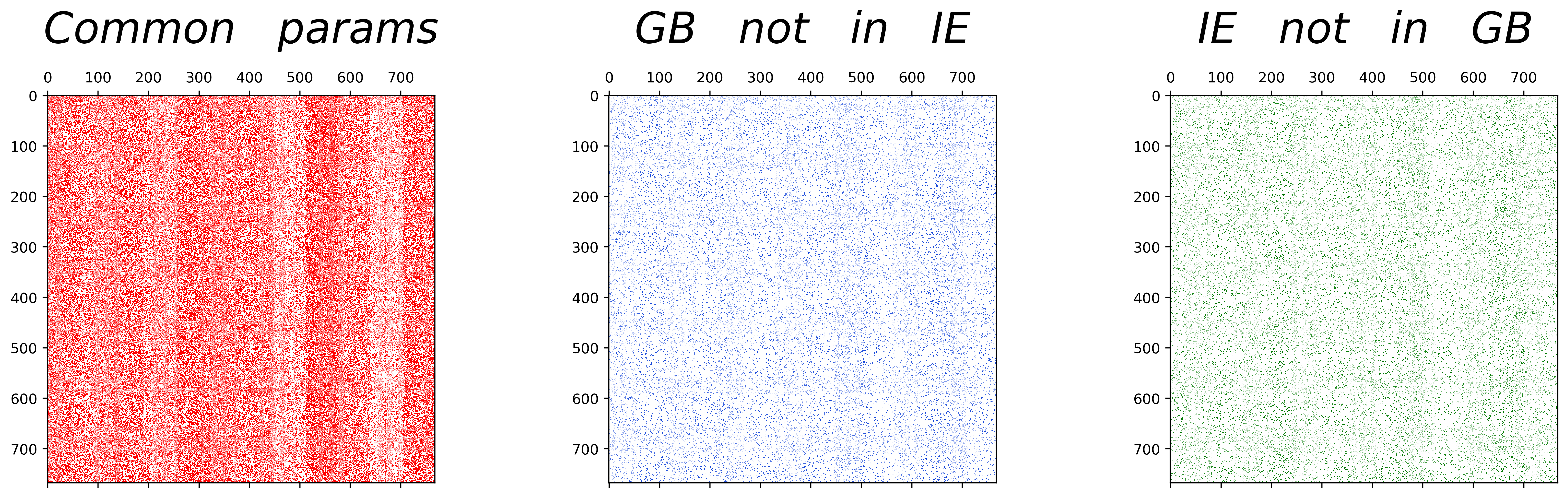}
    \caption{Comparison of the optimal subnetworks of two DeBERTa models (layer 0, attention output matrix) trained on British (GB) and Irish (IE) linguistic variation, respectively. The matrix on the left shows the number of common parameters between the two matrices (subnetwork overlap), while the middle one shows the location of the optimal parameters of the GB subnetwork not present in IE, and on the right the exact opposite. Blank values refer to the values not belonging to the optimal network and thus the collection of points that the KEN algorithm has reset to their pre-training value. Additional results are shown in Apx. \ref{apx: more results}}
    \label{fig: tre tabelle}
\end{figure*}

This section introduces the core components of our research: the EPIC corpus and the KEN pruning algorithm. Sec. \ref{subsec: EPIC} provides an in-depth exploration of the EPIC corpus, explaining its composition and the diverse language varieties it encompasses. Sec. \ref{subsec: KEN} analyzes the KEN pruning algorithm, emphasizing its key role in transformer model optimization.

\subsection{EPIC Corpus} \label{subsec: EPIC}
The EPIC \cite{frenda-etal-2023-epic} corpus consists of 3,000 conversations from social media platforms. It covers five different varieties of English, including Australian (AU), British (GB), Irish (IE), Indian (IN) and American (US). The corpus offers valuable insights into how cultural and linguistic factors shape the perception of irony, giving a comprehensive analysis of it from different perspectives. 

To ensure the authenticity of the data, EPIC sources its content from Twitter and Reddit, capturing informal communication across different regions and demographic areas. Rigorous data curation guarantees the inclusion of potential ironies while maintaining a balanced distribution across language varieties, mitigating selection bias.
Native speakers from each country independently label instances as ironic or non-ironic, using a multi-perspective annotation process. This ensures a robust and nuanced understanding of cultural humor. Annotators possess robust language skills and familiarity with online communication styles, reinforcing the reliability of their judgments. The inclusive approach in both data collection and annotation facilitates the development of \textit{perspective-aware} models \cite{akhtar2021whose} that account for cultural and linguistic variations.

\subsection{KEN algorithm} \label{subsec: KEN}
KEN (\textbf{K}ernel density \textbf{E}stimator for \textbf{N}eural network compression) \cite{mastromattei2024less}, is a pruning algorithm designed to extract the most essential subnetwork from transformer models. It exploits the \textit{winning ticket lottery hypothesis} \cite{frankle2018lottery}, according to which an optimal subset of fine-tuned parameters maintains the same performance as the original one.

KEN leverages Kernel Density Estimations (KDEs) to generalize point distributions for each row of a transformer matrix, resulting in a streamlined version of the original fine-tuned model. By pinpointing the $k$ most representative parameters within each distribution, KEN effectively prunes the network, preserving them while reverting the remaining parameters to their pre-trained state.
KEN archives minimum parameter reduction between 25\% and 60\% for specific models, maintaining equivalent or better performance than their unpruned counterparts. 
The resultant subnetwork can be seamlessly archived and reintegrated into its pre-trained configuration for diverse downstream applications. This approach not only significantly reduces model size but also enhances efficiency and flexibility across various tasks.

\begin{table*}[h!]
\begin{adjustbox}{width=\linewidth}
\subfloat[Pertentage of parameter reset after the KEN pruning step for all the models on each language variation subsets analyzed. The percentage indicates the number of parameters reset to their pre-trained value in the entire model]{
\begin{tabular}{lccccc}
Model      & AU    & GB    & IE    & IN    & US    \\ \hline
Bert       & 47.54 & 58.03 & 58.03 & 58.03 & 58.03 \\ \hline
DistilBert & 56.26 & 34.39 & 50.79 & 50.79 & 56.26 \\ \hline
DeBerta    & 44.88 & 55.91 & 55.91 & 55.91 & 55.91 \\ \hline
Ernie      & 58.03 & 47.54 & 58.03 & 58.03 & 58.03  \\ \hline
Electra    & 91.18 & 91.18 & 64.75 & 91.18 & 82.37
\end{tabular} \label{tab: Parameter reset}
}
\quad
\subfloat[Variation of the F1-weighted measure across all the language variation subsets after the KEN pruning step. Positive values indicate a score improvement compared to the unpruned version]{
\begin{tabular}{lccccc}
Model      & AU   & GB   & IE   & IN   & US   \\ \hline
Bert       & +2.0 & +2.1 & +5.5 & +4.6 & +0.0 \\ \hline
DistilBert & +0.6 & +0.0 & +3.5 & +2.4 & +0.0 \\ \hline
DeBerta    & +1.3 & +2.9 & +7.2 & +1.4 & +0.0 \\ \hline
Ernie      & +0.0 & +0.0 & +0.0 & +13.5 & +0.0 \\ \hline
Electra    & +5.2 & +0.7 & +1.5 & +0.1 & +2.1
\end{tabular}  \label{tab: Perfomance}
}
\end{adjustbox}
\caption{Result obtained during our experiment: Tab. \ref{tab: Parameter reset} shows the percentage of parameter reset of each model in all language variation subsets analyzed while Tab. \ref{tab: Perfomance} presents per F1-weighted performance variation obtained.}
\label{tab: results}
\end{table*}
\section{Results} \label{sec:Results}

The KEN algorithm is an effective method for selecting the best model parameters for each language variation. The rate at which these parameters are reset varies across different architectures, as shown in Tab. \ref{tab: Parameter reset}. However, this resetting rate consistently exceeds 50\% on average. Surprisingly, despite the substantial resetting, performance actually improves in most cases, as demonstrated by the F1-weighted scores in Tab. \ref{tab: Perfomance}. Notably, these results were achieved through tuning steps on relatively small data sets, with only 600 examples per variation. It is essential to note that our primary goal was not to establish new state-of-the-art (SoTa) models, but rather to investigate the impact of language variations on model parameters within each architecture examined. From this perspective, the results are encouraging and demonstrate a positive impact. Additionally, the varying percentages of parameter resets among linguistic variations using the same architecture contribute to a more nuanced understanding of the optimal subnetworks and their comparison.

After examining subnetwork structures, it was discovered that two optimal subnetworks share at least 60\% of their parameters. This percentage, however, does not take into account parameters reset by KEN, which could significantly impact the final result. Tab. \ref{tab: overlap} indicates that Indian (IN) and American (US) variations have the highest overlap, with more than 90\% in three out of five models. British (GB) and Irish (IE) also have considerable overlap across all models, which is highly desirable. Despite extensive analysis, identifying the most distinct variants remains challenging, as the percentage difference between pairs of language variations across all models is relatively small.

\begin{table}[t!]
\centering
\begin{tabular}{ll|ccccc}
Subnet A & Subnet B & BERT            & DeBERTa         & DistilBERT & Ernie           & Electra \\ \hline
AU       & GB       & 69.73           & 69.94           & 61.69      & 69.81           & 89.49   \\
AU       & IE       & 69.79           & 69.94           & 75.22      & 82.72           & 23.15   \\
AU       & IN       & 69.73           & 69.94           & 75.17      & 83.22           & 87.6    \\
AU       & US       & 69.73           & 69.94           & 83.42      & 83.22           & 29.09   \\ \hline
GB       & IE       & 83.02           & 82.74           & 69.38      & 69.76           & 23.15   \\
GB       & IN       & 82.59           & 82.71           & 69.38      & 69.81           & 86.95   \\
GB       & US       & 82.59           & 82.71           & 61.66      & 69.81           & 29.06   \\ \hline
IE       & IN       & 82.6            & 82.86           & 85.85      & 82.39           & 23.15   \\
IE       & US       & 82.6            & 82.86           & 75.17      & 82.39           & 69.68   \\ \hline
IN       & US       & \textgreater 90.0 & \textgreater 90.0 & 76.22      & \textgreater 90.0 & 29.45  
\end{tabular}
\caption{Similarity percentages between subnetworks specific to language variation. Percentages are obtained by comparing for each model the number of non-reset parameters within each attention (or similarity) layers}
\label{tab: overlap}
\end{table}

In addition to tabular descriptions, we have graphically presented the results obtained. Through $KEN_{viz}$, three different types of results are visualized: (1) the subnetwork overlap of two language variations within the same selected matrix layer, (2) fine-tuned parameters chosen for the linguistic variation A but not for B and (3) the reverse. Fig. \ref{fig: tre tabelle} showcases one of the obtained results, while Apx. \ref{apx: more results} provides more case studies by analyzing results across all models in their last attention layer for specific linguistic variations. These graphical representations offer insights into the precise placement of optimal parameters and the shared or differing structures between models.

\section{Conclusion} \label{sec: conclusion}

This study conducted a thorough analysis of different transformer models to discover their divergences in detecting irony when trained on different linguistic variants. We uesd the EPIC corpus and created language-variant-specific datasets for five English variations (American, British, Indian, Irish and Australian). Using the KEN pruning algorithm, we extracted optimal subnetworks from five transformer architectures (BERT, DistilBERT, DeBERTa, Ernie and Electra) tailored to each language variation.
Our study revealed that different linguistic variations share a remarkable number of parameters, regardless of the architecture used. We provided insights into the similarity of each pair of optimized subnetwork linguistic variations by reporting the percentage of common parameters. However, we found it challenging to rank the dissimilarity since the shared parameter percentage remained consistently high in all cases.
To enhance our understanding of how linguistic diversity manifests in the models, we used $KEN_{viz}$ to provide a graphical view of the specific locations of shared and distinct parameters across models and language variations.

Although there are limitations such as the size of the dataset, our study demonstrates that training transformer models and adapting them to linguistic variations yield highly similar output models demonstrating how their difference is intrinsic to their parameter values.

\newpage
\bibliographystyle{unsrt}  
\bibliography{references} 
\newpage
\appendix
\newpage
\section{$KEN_{viz}$ outputs} \label{apx: more results}
In this appendix, we present some graphical results obtained using $KEN_{viz}$ by analyzing the output of attention matrices in the first and last levels for each model analyzed. We selected several pairs of linguistic variations for each model that showed the most interesting results based on the findings in Tab.\ref{tab: overlap}. These visual results highlight the commonalities found within the optimal subnetworks and show the difficulty of finding differences between them. However, we can observe that in some cases, parameter selection focuses more on certain areas than others.

\subsection{Results on BERT model}
\begin{figure}[h!]
    \begin{subfigure}[h]{0.48\linewidth}
            \centering
            \includegraphics[width=\linewidth]{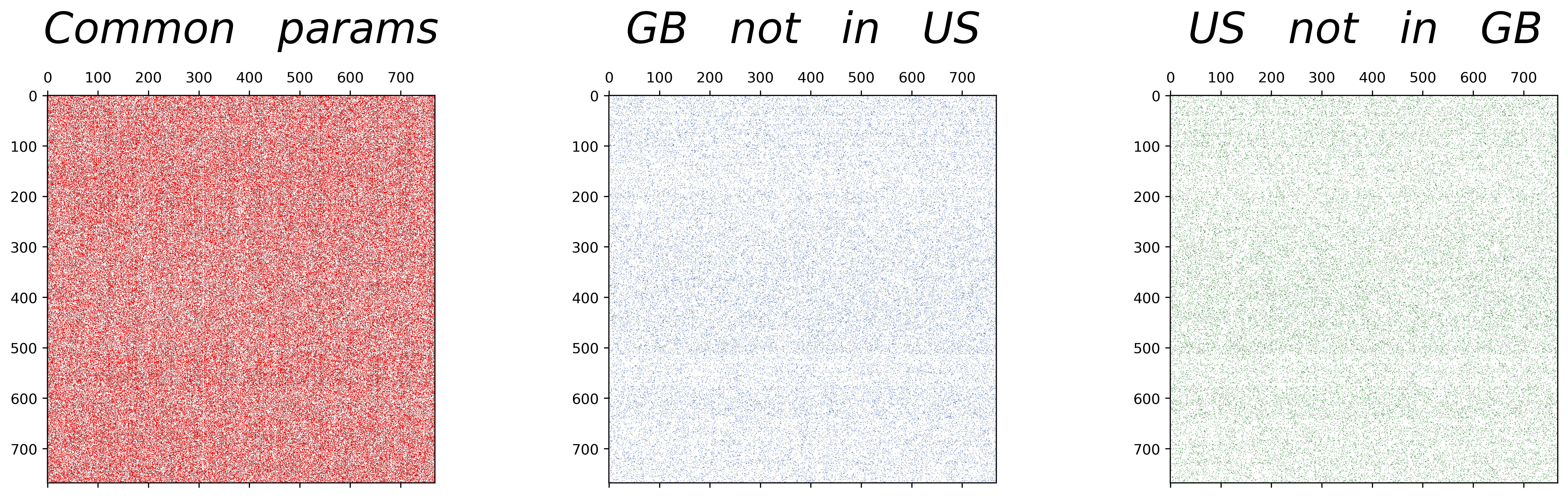}
            \caption{Key matrices}
        \end{subfigure}
        \hfill
        \begin{subfigure}[h]{0.48\linewidth}
            \centering
            \includegraphics[width=\linewidth]{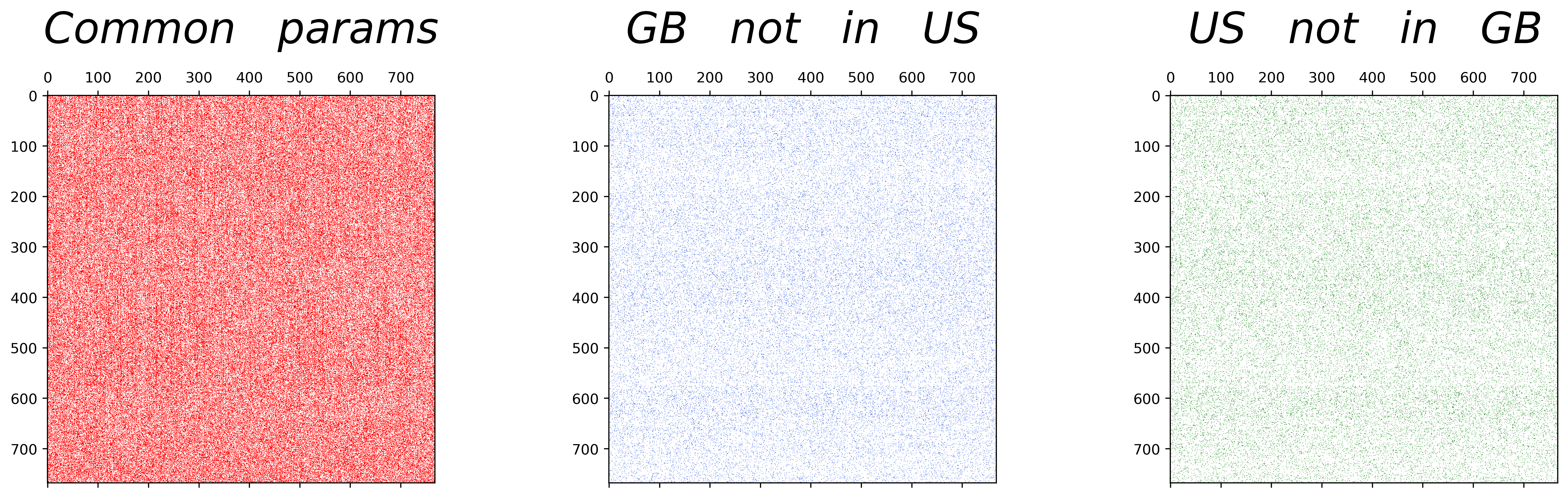}
            \caption{Query matrices}
        \end{subfigure}
        \hfill
        \begin{subfigure}[h]{0.48\linewidth}
            \centering
            \includegraphics[width=\linewidth]{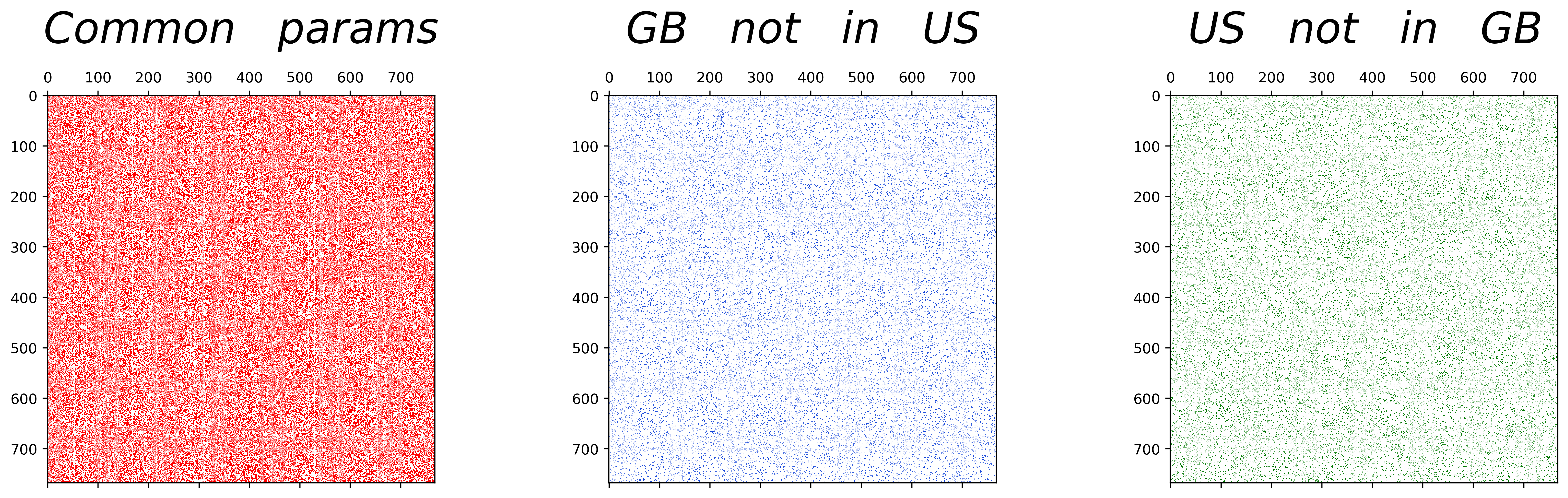}
            \caption{Value matrices}
        \end{subfigure}
    \caption{Layer 0 attention matrices}
\end{figure}

\begin{figure}[h!]
    \begin{subfigure}[h]{0.48\linewidth}
            \centering
            \includegraphics[width=\linewidth]{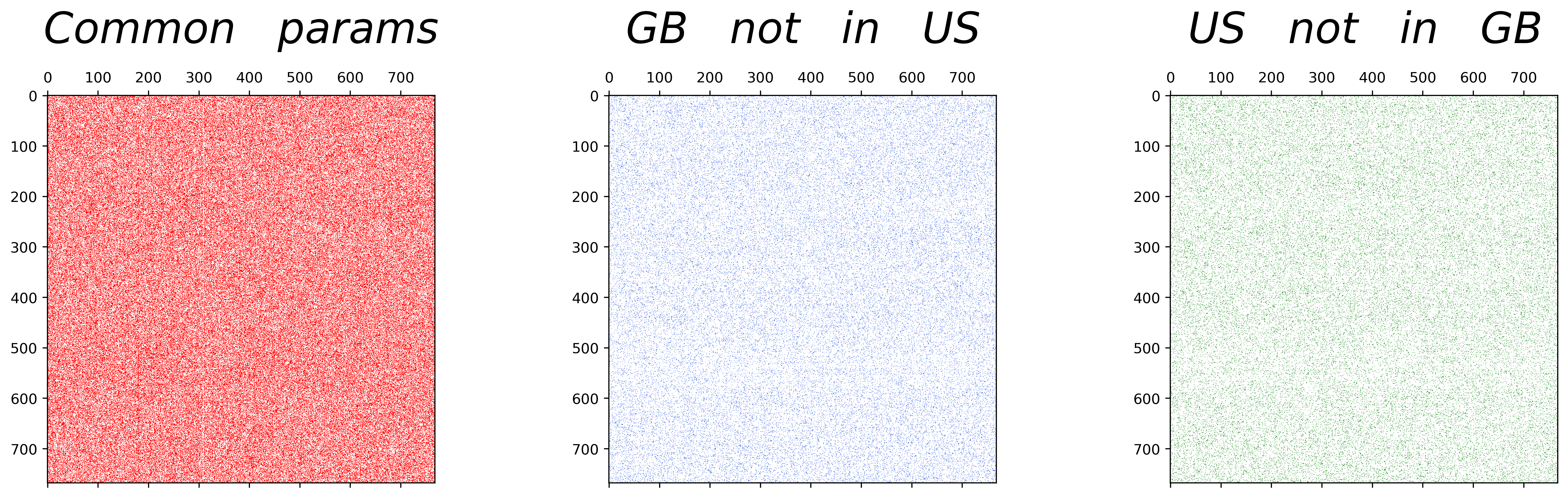}
            \caption{Key matrices}
        \end{subfigure}
        \hfill
        \begin{subfigure}[h]{0.48\linewidth}
            \centering
            \includegraphics[width=\linewidth]{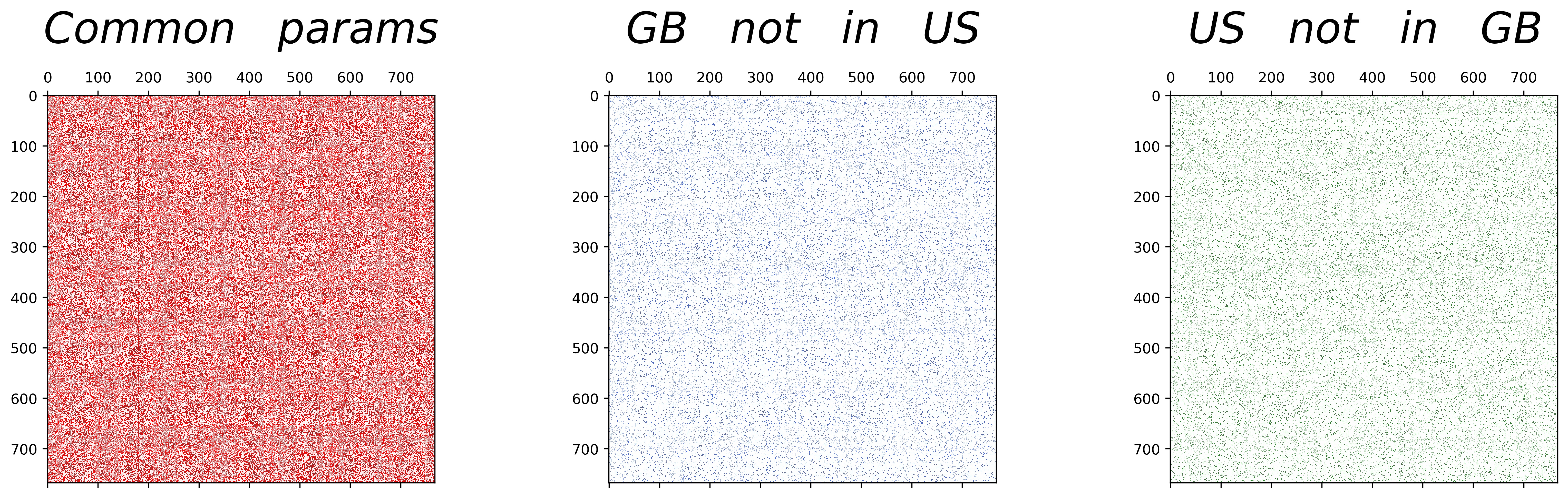}
            \caption{Query matrices}
        \end{subfigure}
        \hfill
        \begin{subfigure}[h]{0.48\linewidth}
            \centering
            \includegraphics[width=\linewidth]{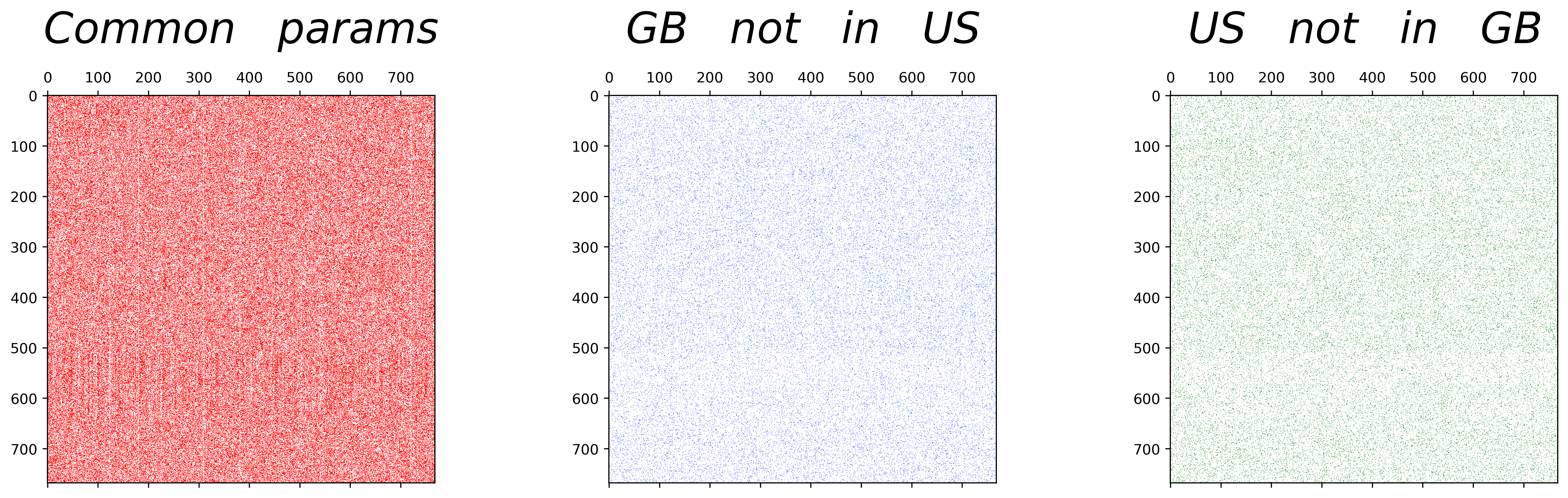}
            \caption{Value matrices}
        \end{subfigure}
    \caption{Layer 12 attention matrices}
\end{figure}
\newpage

\subsection{Results on DistilBERT model}
\begin{figure}[h!]
    \begin{subfigure}[h]{0.48\linewidth}
            \centering
            \includegraphics[width=\linewidth]{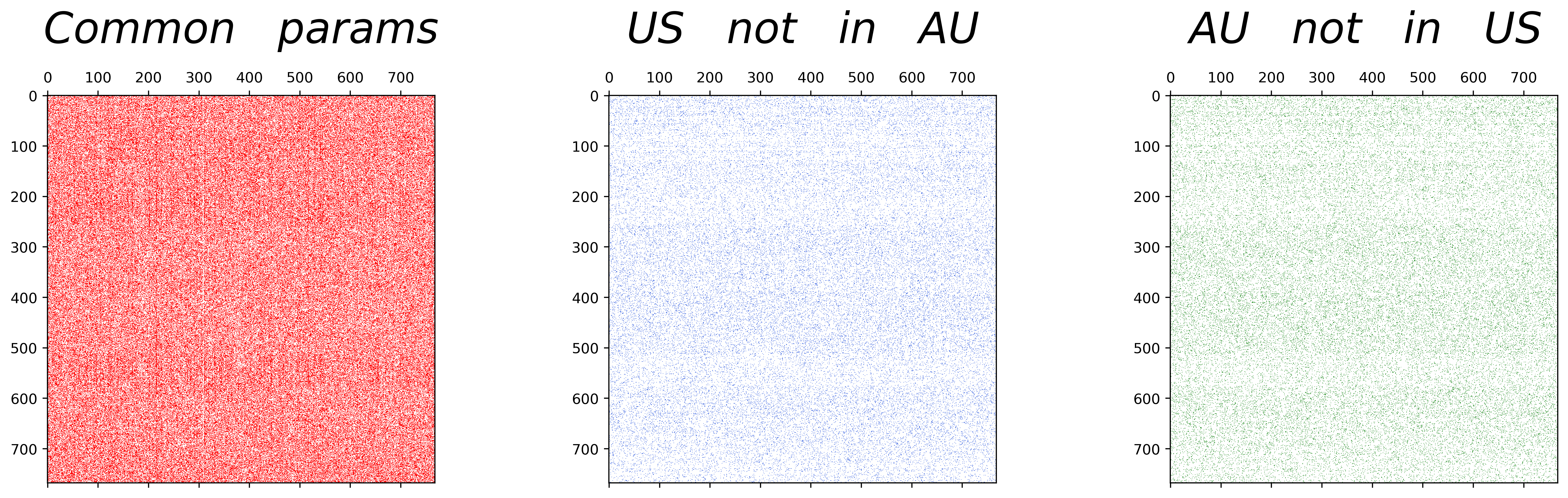}
            \caption{\texttt{k\_lin} matrices}
        \end{subfigure}
        \hfill
        \begin{subfigure}[h]{0.48\linewidth}
            \centering
            \includegraphics[width=\linewidth]{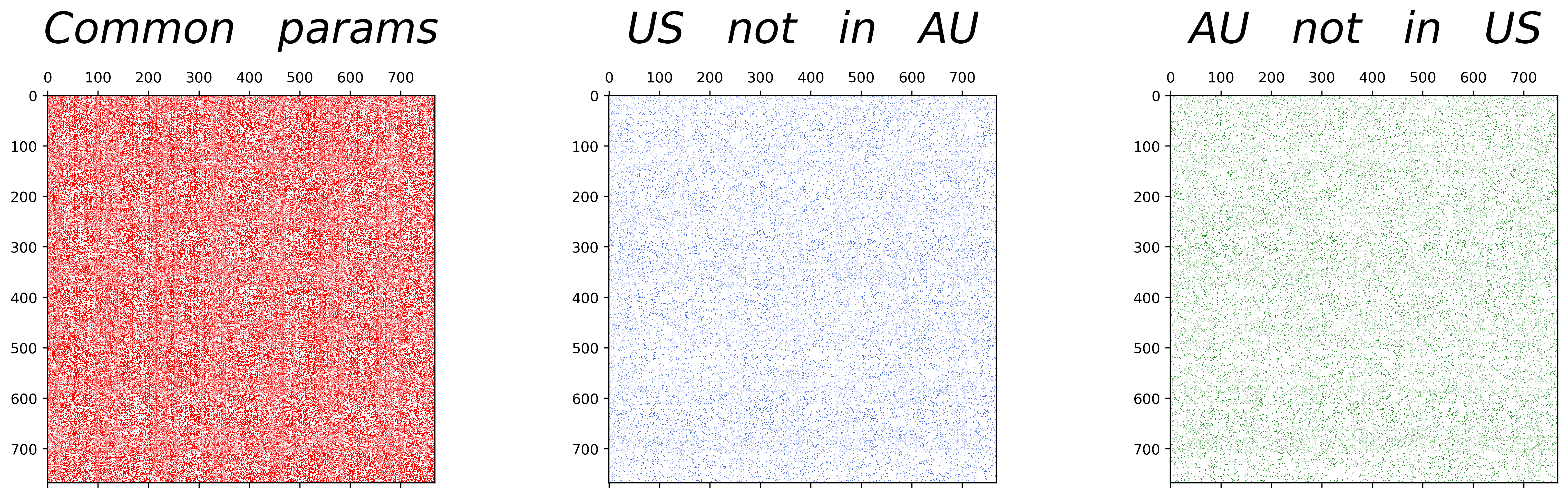}
            \caption{\texttt{q\_lin} matrices}
        \end{subfigure}
        \hfill
        \begin{subfigure}[h]{0.48\linewidth}
            \centering
            \includegraphics[width=\linewidth]{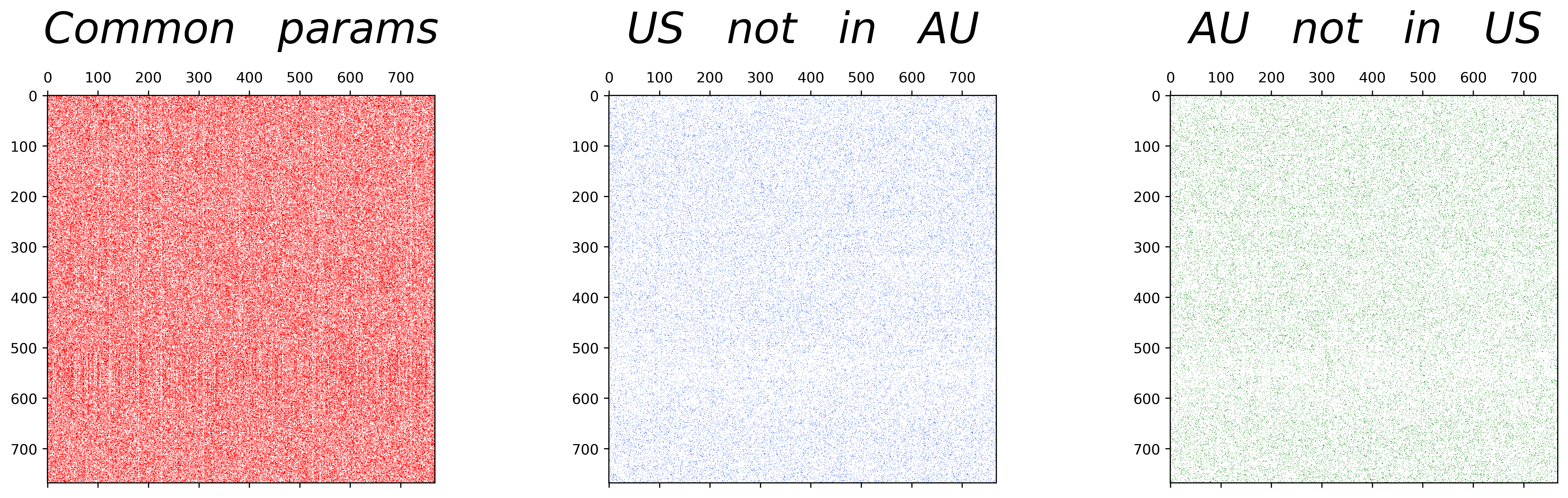}
            \caption{\texttt{v\_lin} matrices}
        \end{subfigure}
    \caption{Layer 0 attention matrices}
\end{figure}

\begin{figure}[h!]
    \begin{subfigure}[h]{0.48\linewidth}
            \centering
            \includegraphics[width=\linewidth]{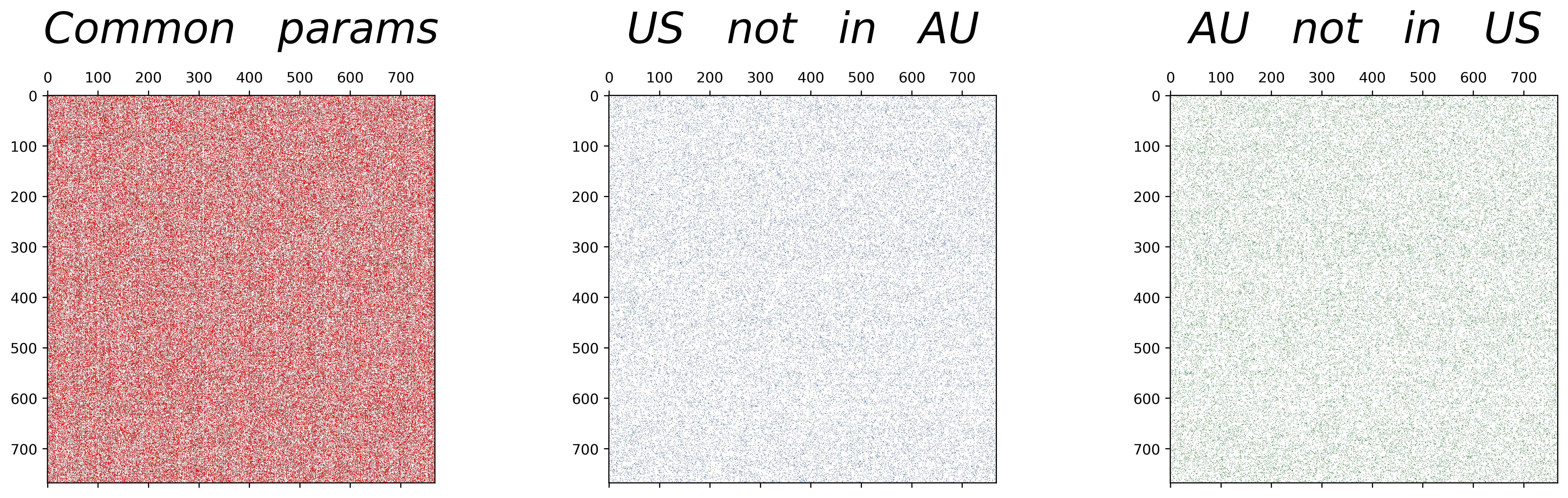}
            \caption{\texttt{k\_lin} matrices}
        \end{subfigure}
        \hfill
        \begin{subfigure}[h]{0.48\linewidth}
            \centering
            \includegraphics[width=\linewidth]{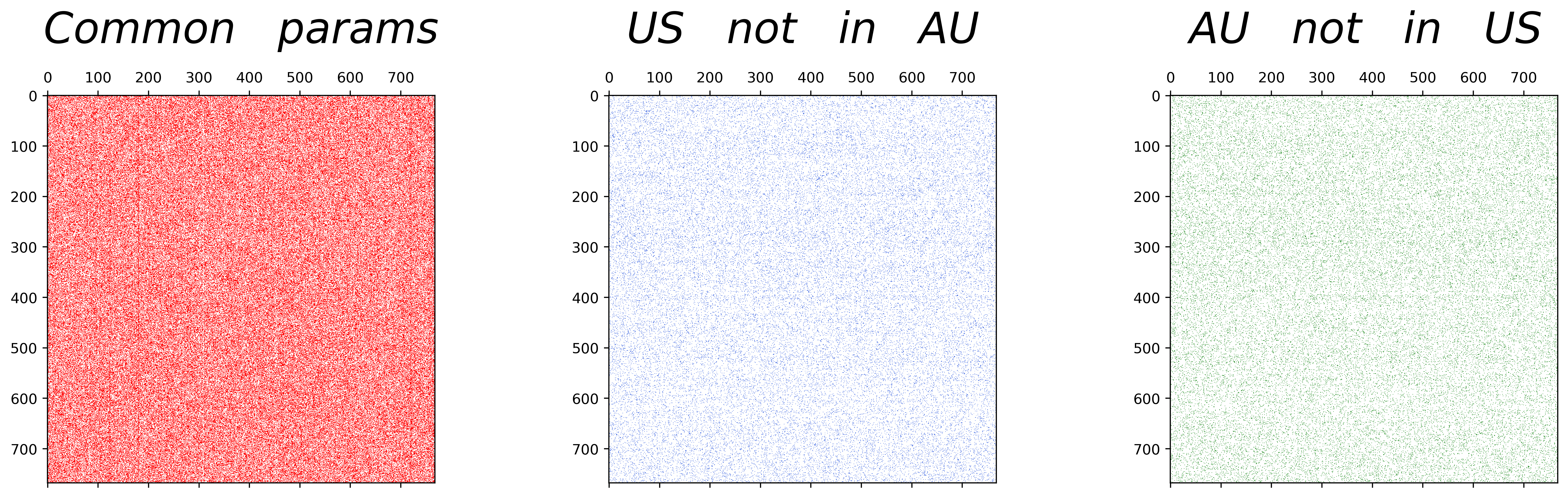}
            \caption{\texttt{q\_lin} matrices}
        \end{subfigure}
        \hfill
        \begin{subfigure}[h]{0.48\linewidth}
            \centering
            \includegraphics[width=\linewidth]{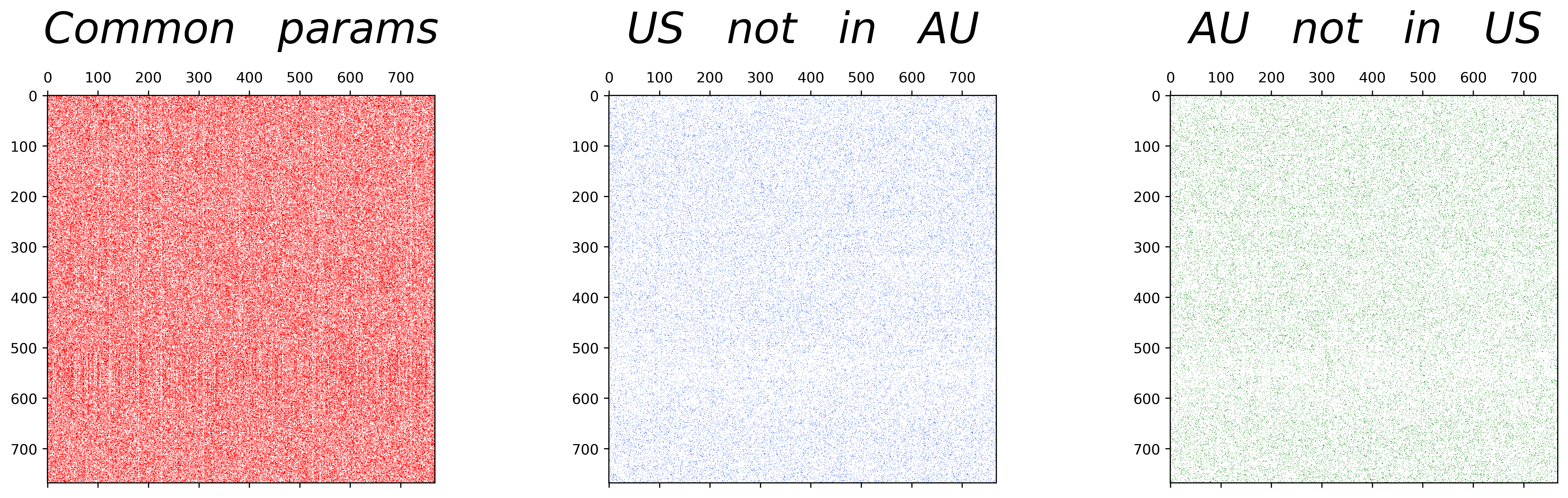}
            \caption{\texttt{v\_lin} matrices}
        \end{subfigure}
    \caption{Layer 5 attention matrices}
\end{figure}
\newpage

\subsection{Results in DeBerta model}
\begin{figure}[h!]
    \begin{subfigure}[h!]{0.48\linewidth}
            \centering
            \includegraphics[width=\linewidth]{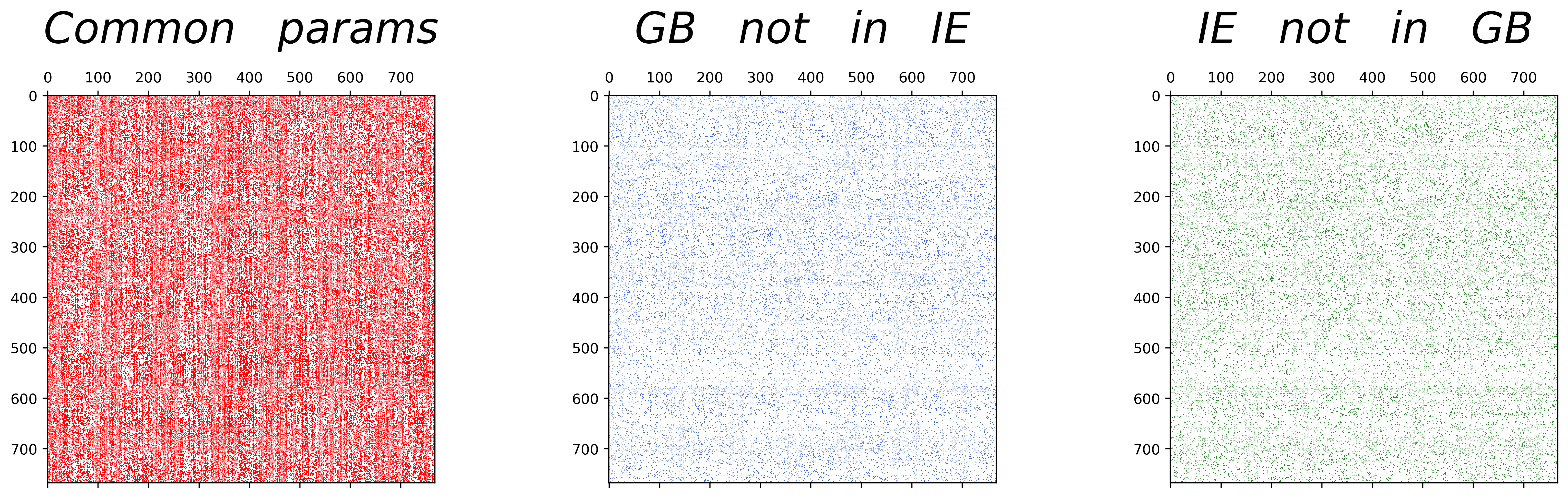}
            \caption{\texttt{q\_proj} matrices}
        \end{subfigure}
        \hfill
        \begin{subfigure}[h]{0.48\linewidth}
            \centering
            \includegraphics[width=\linewidth]{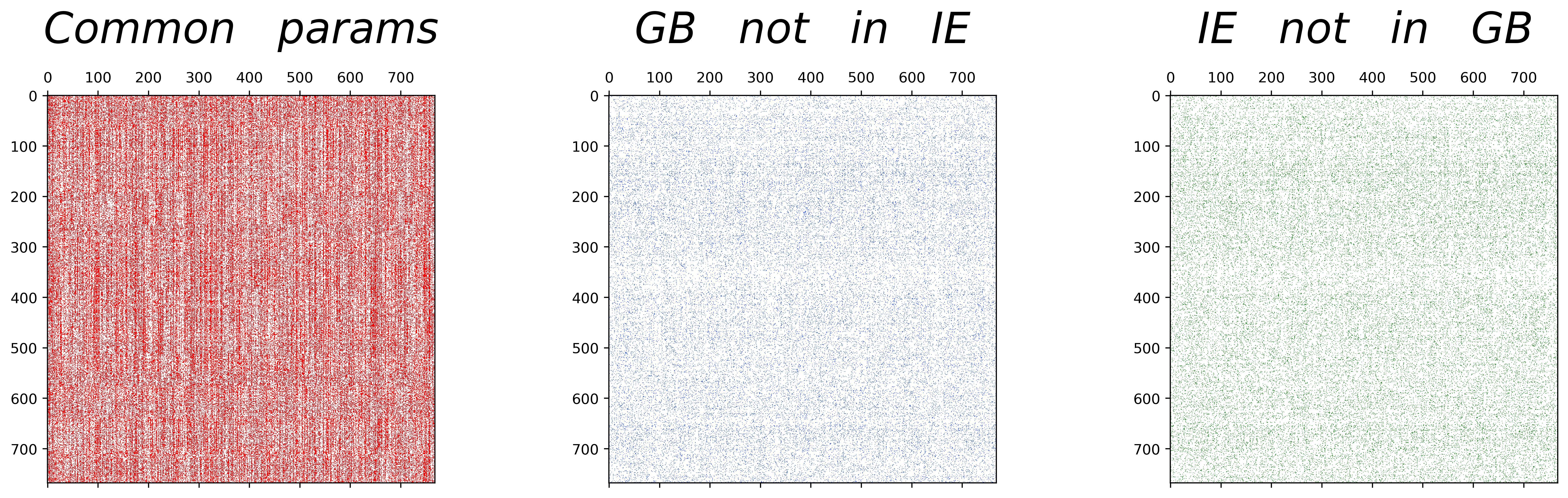}
            \caption{\texttt{pos\_proj} matrices}
        \end{subfigure}
        \hfill
        \begin{subfigure}[h]{0.48\linewidth}
            \centering
            \includegraphics[width=\linewidth]{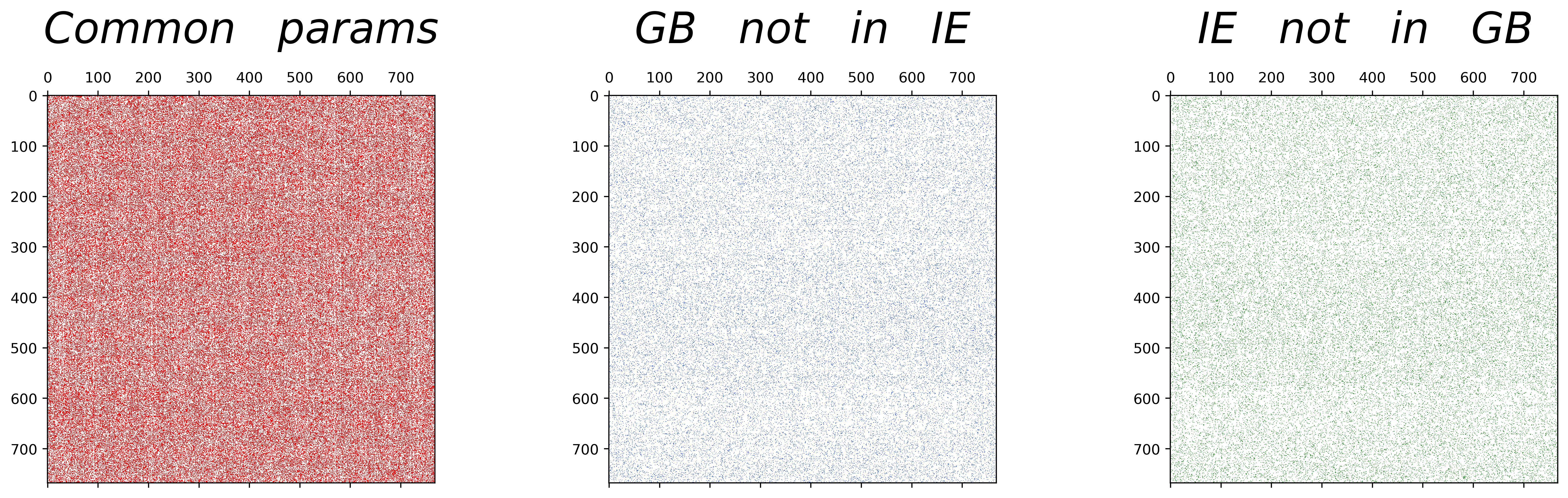}
            \caption{\texttt{in\_proj} matrices}
        \end{subfigure}
        \hfill
        \begin{subfigure}[h]{0.48\linewidth}
            \centering
            \includegraphics[width=\linewidth]{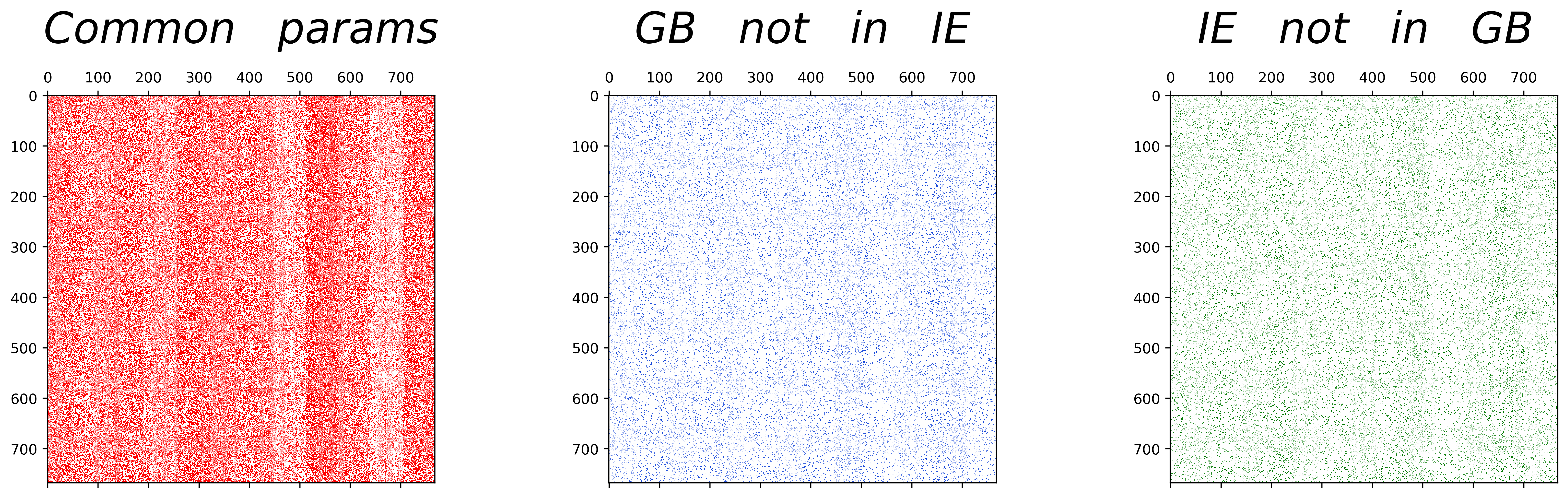}
            \caption{Output matrices}
        \end{subfigure}
    \caption{Layer 0 attention matrices}
\end{figure}

\begin{figure}[h!]
    \begin{subfigure}[h!]{0.48\linewidth}
            \centering
            \includegraphics[width=\linewidth]{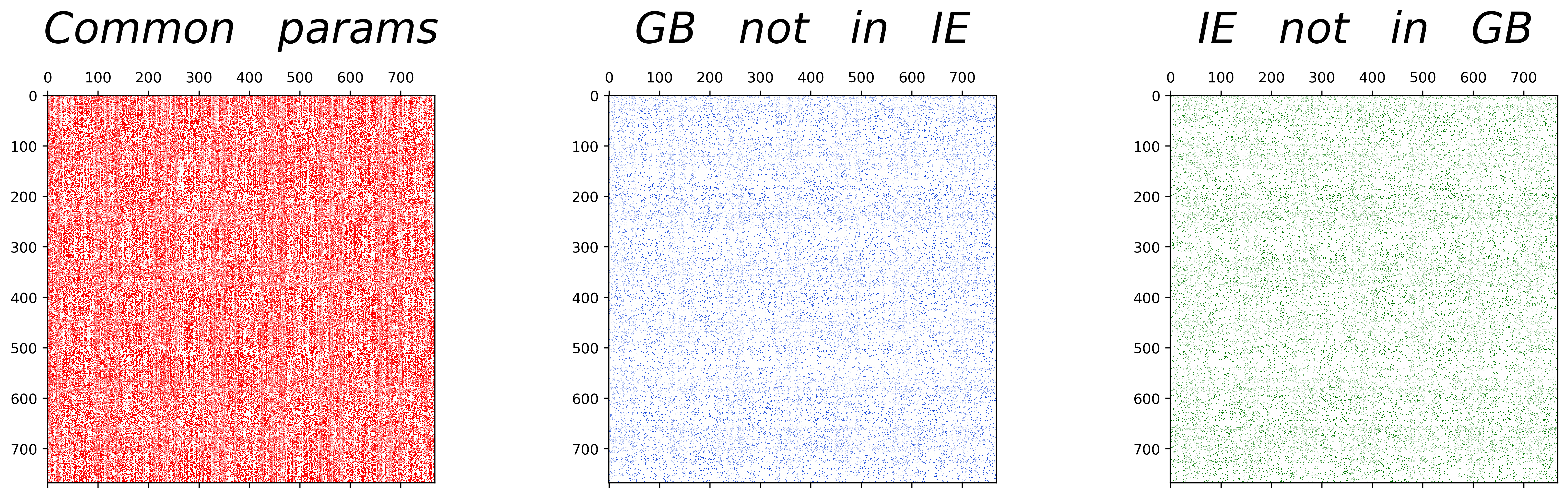}
            \caption{\texttt{q\_proj} matrices}
        \end{subfigure}
        \hfill
        \begin{subfigure}[h]{0.48\linewidth}
            \centering
            \includegraphics[width=\linewidth]{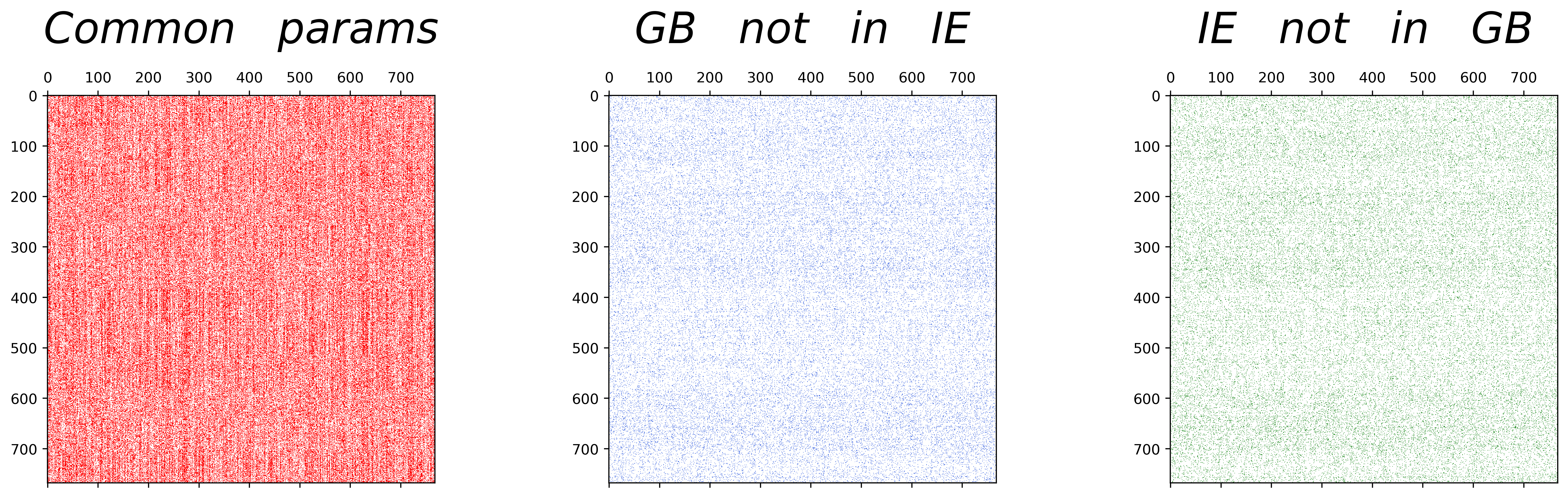}
            \caption{\texttt{pos\_proj} matrices}
        \end{subfigure}
        \hfill
        \begin{subfigure}[h]{0.48\linewidth}
            \centering
            \includegraphics[width=\linewidth]{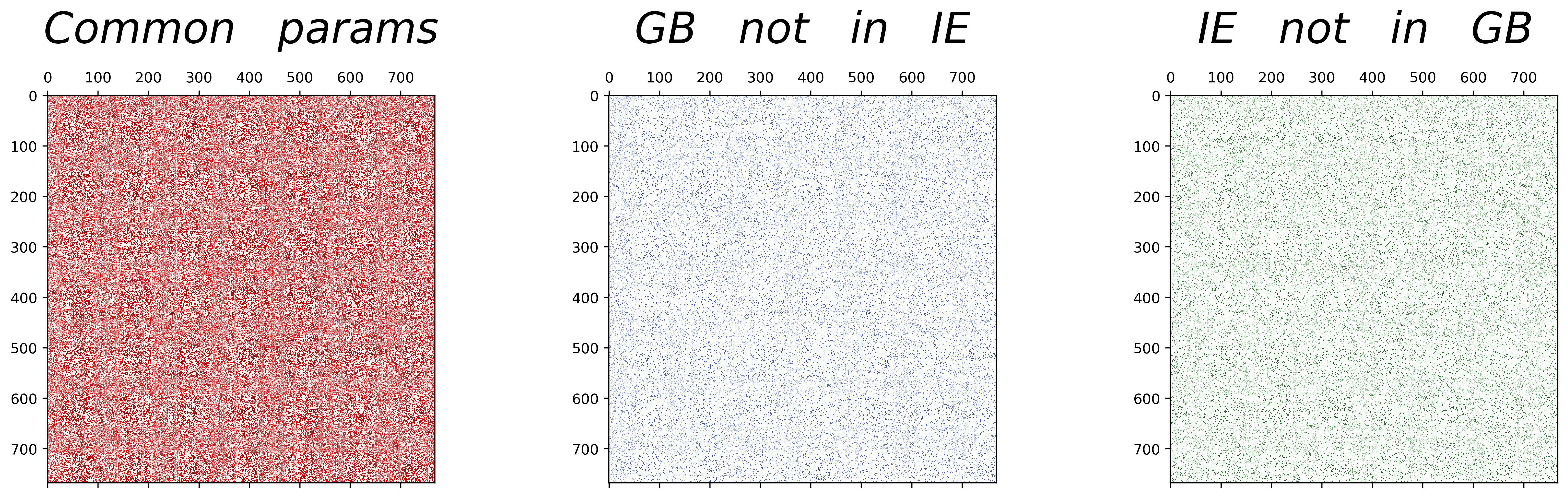}
            \caption{\texttt{in\_proj} matrices}
        \end{subfigure}
        \hfill
        \begin{subfigure}[h]{0.48\linewidth}
            \centering
            \includegraphics[width=\linewidth]{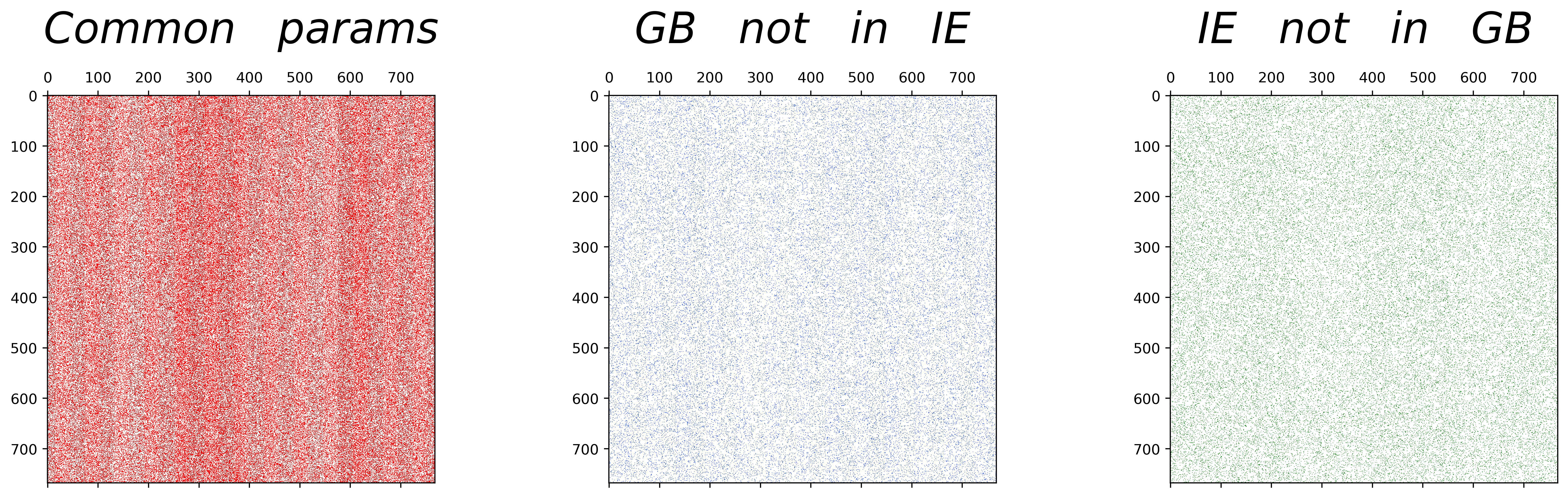}
            \caption{Output matrices}
        \end{subfigure}
    \caption{Layer 12 attention matrices}
\end{figure}
\newpage

\subsection{Results on Ernie model}
\begin{figure}[h!]
    \begin{subfigure}[h]{0.48\linewidth}
            \centering
            \includegraphics[width=\linewidth]{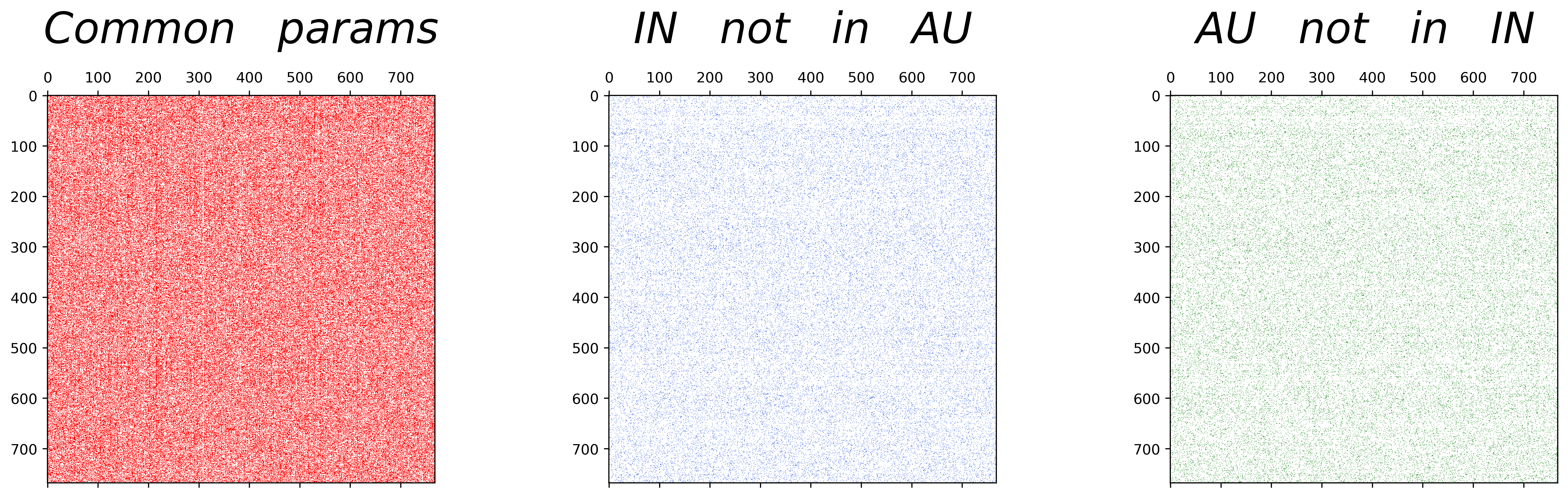}
            \caption{Key matrices}
        \end{subfigure}
        \hfill
        \begin{subfigure}[h]{0.48\linewidth}
            \centering
            \includegraphics[width=\linewidth]{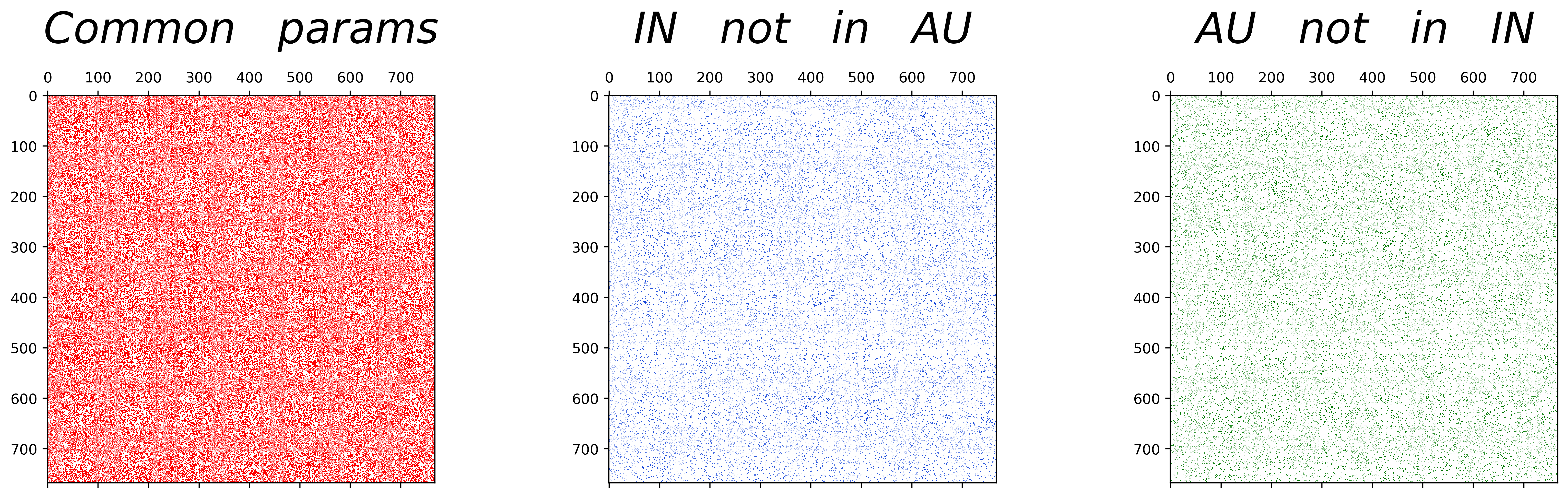}
            \caption{Query matrices}
        \end{subfigure}
        \hfill
        \begin{subfigure}[h]{0.48\linewidth}
            \centering
            \includegraphics[width=\linewidth]{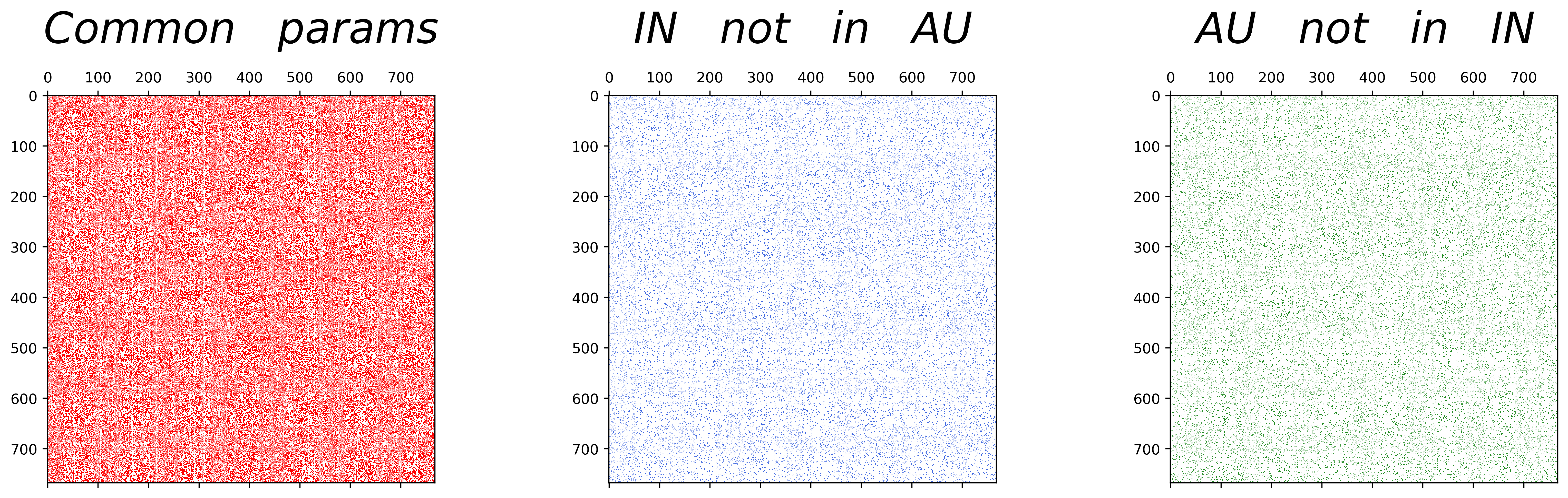}
            \caption{Value matrices}
        \end{subfigure}
        \hfill
        \begin{subfigure}[h]{0.48\linewidth}
            \centering
            \includegraphics[width=\linewidth]{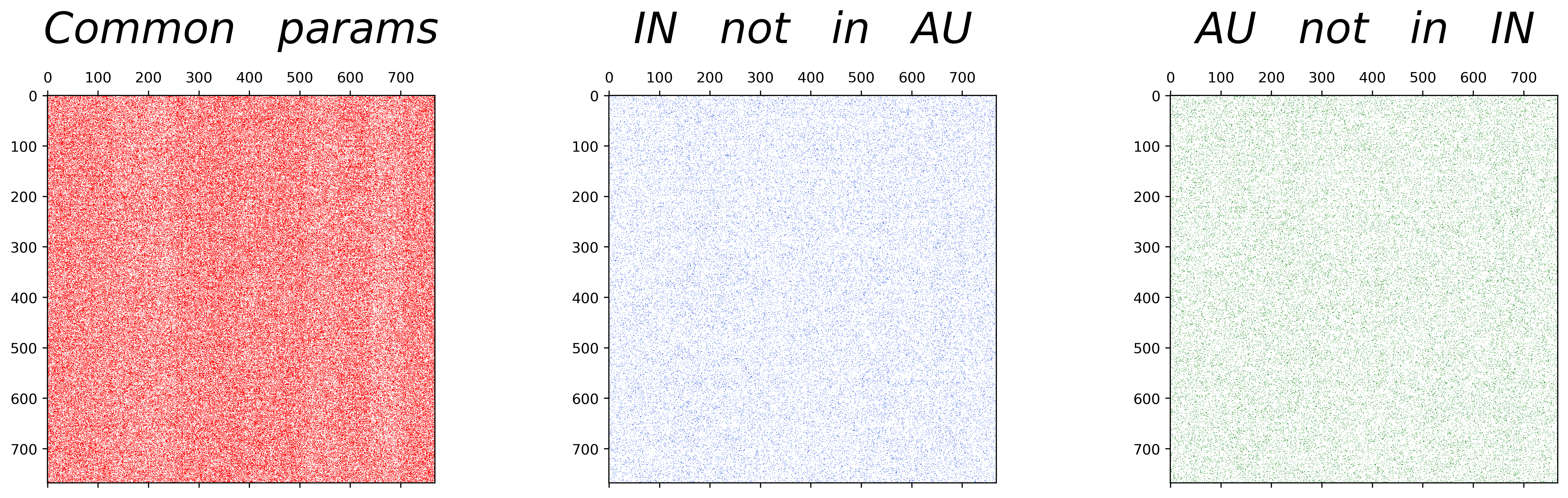}
            \caption{Output matrices}
        \end{subfigure}
    \caption{Layer 0 attention  matrices}
\end{figure}

\begin{figure}[h!]
    \begin{subfigure}[h]{0.48\linewidth}
            \centering
            \includegraphics[width=\linewidth]{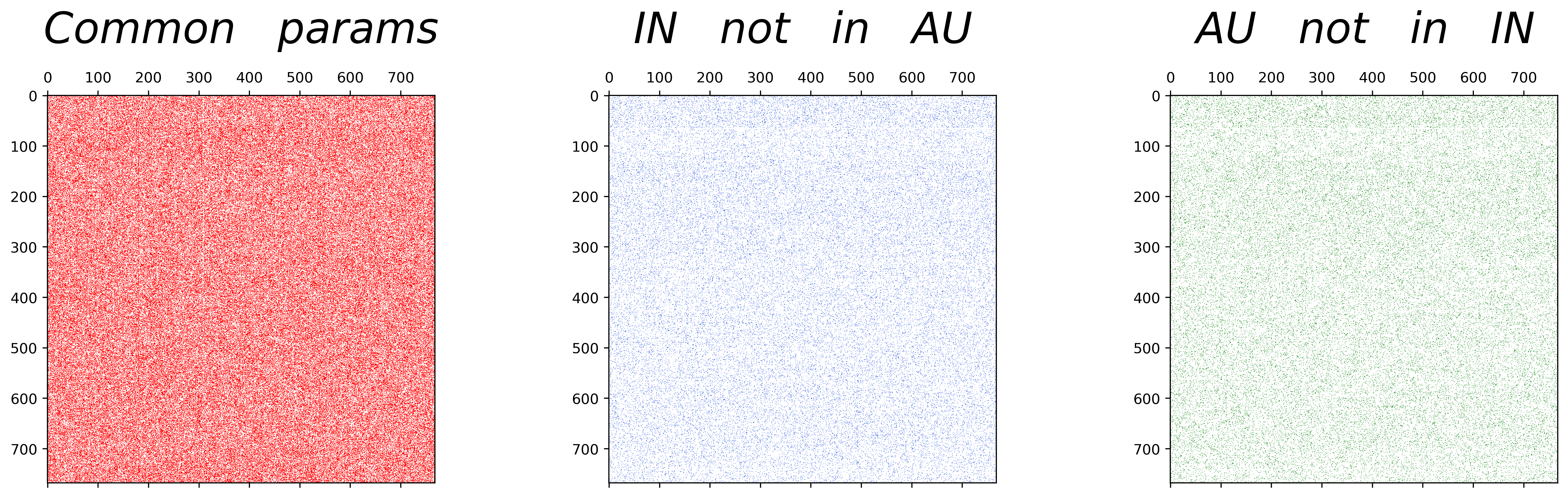}
            \caption{Key matrices}
        \end{subfigure}
        \hfill
        \begin{subfigure}[h]{0.48\linewidth}
            \centering
            \includegraphics[width=\linewidth]{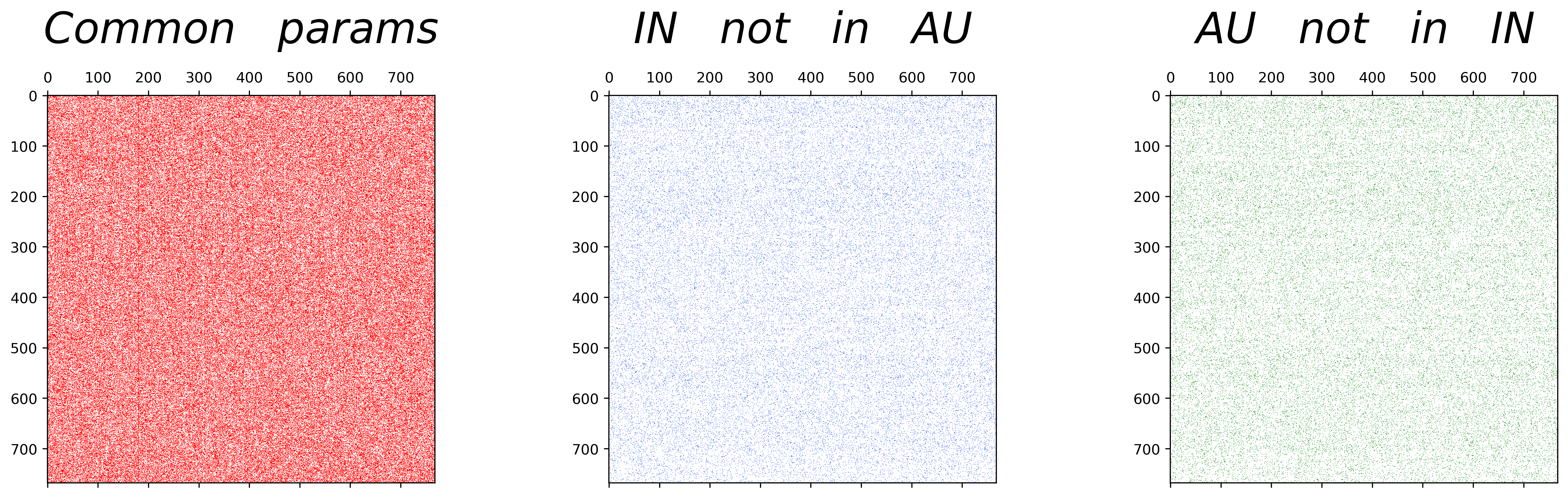}
            \caption{Query matrices}
        \end{subfigure}
        \hfill
        \begin{subfigure}[h]{0.48\linewidth}
            \centering
            \includegraphics[width=\linewidth]{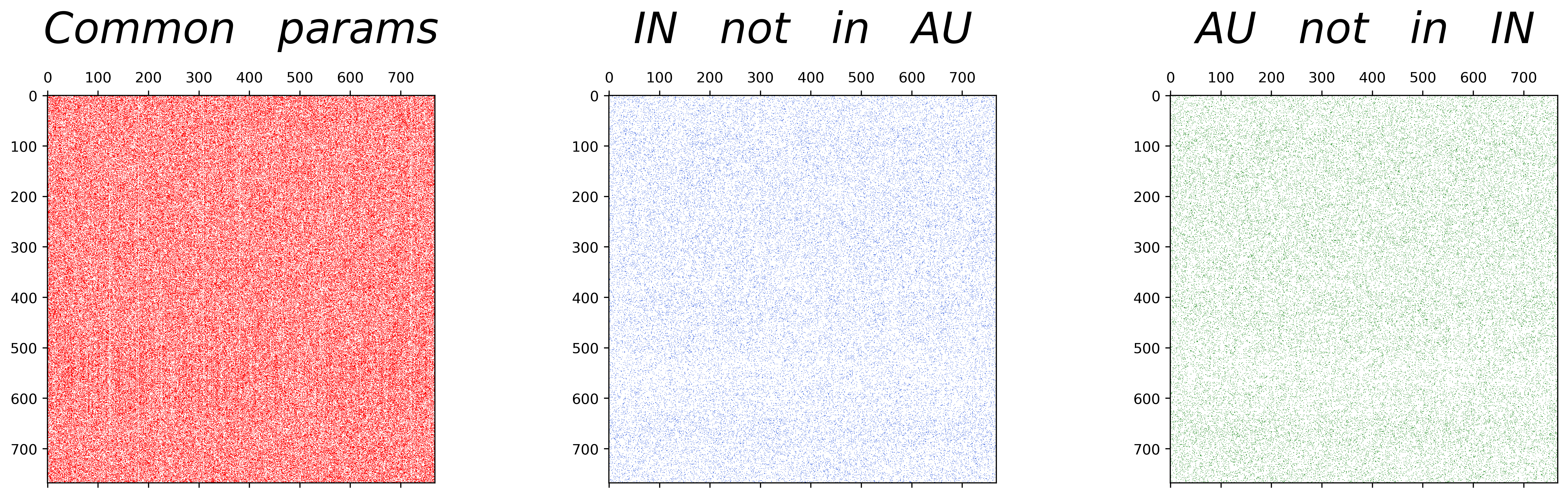}
            \caption{Value matrices}
        \end{subfigure}
        \hfill
        \begin{subfigure}[h]{0.48\linewidth}
            \centering
            \includegraphics[width=\linewidth]{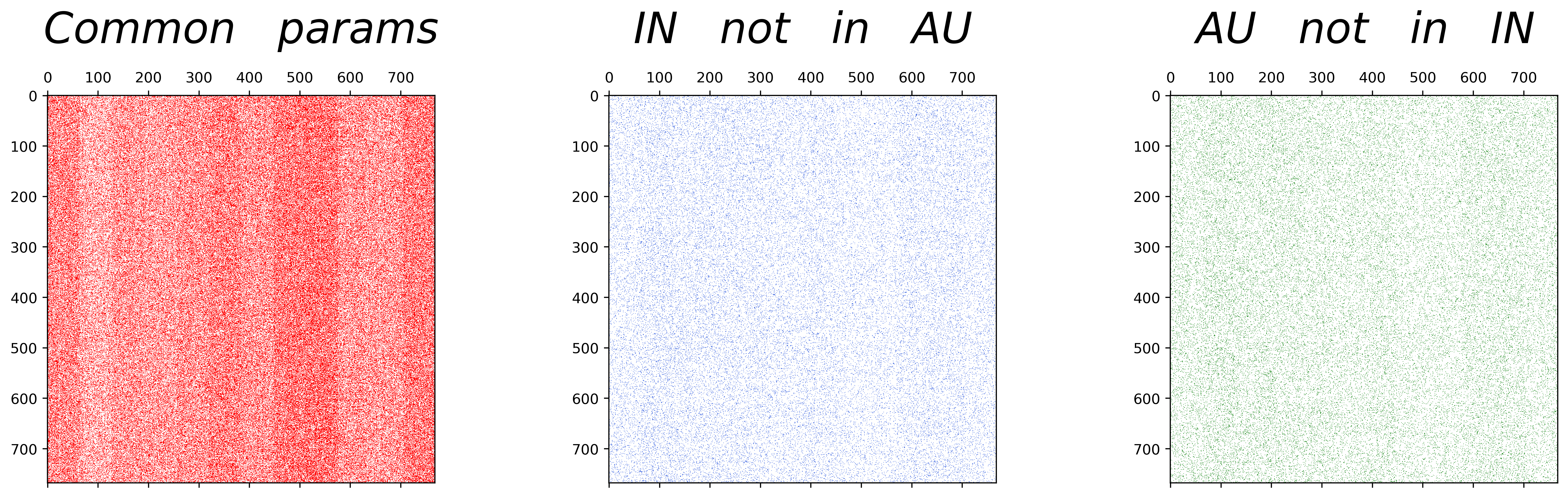}
            \caption{Output matrices}
        \end{subfigure}
    \caption{Layer 11 attention matrices}
\end{figure}
\newpage

\subsection{Results on Electra model}
\begin{figure}[h!]
    \begin{subfigure}[h]{0.48\linewidth}
            \centering
            \includegraphics[width=\linewidth]{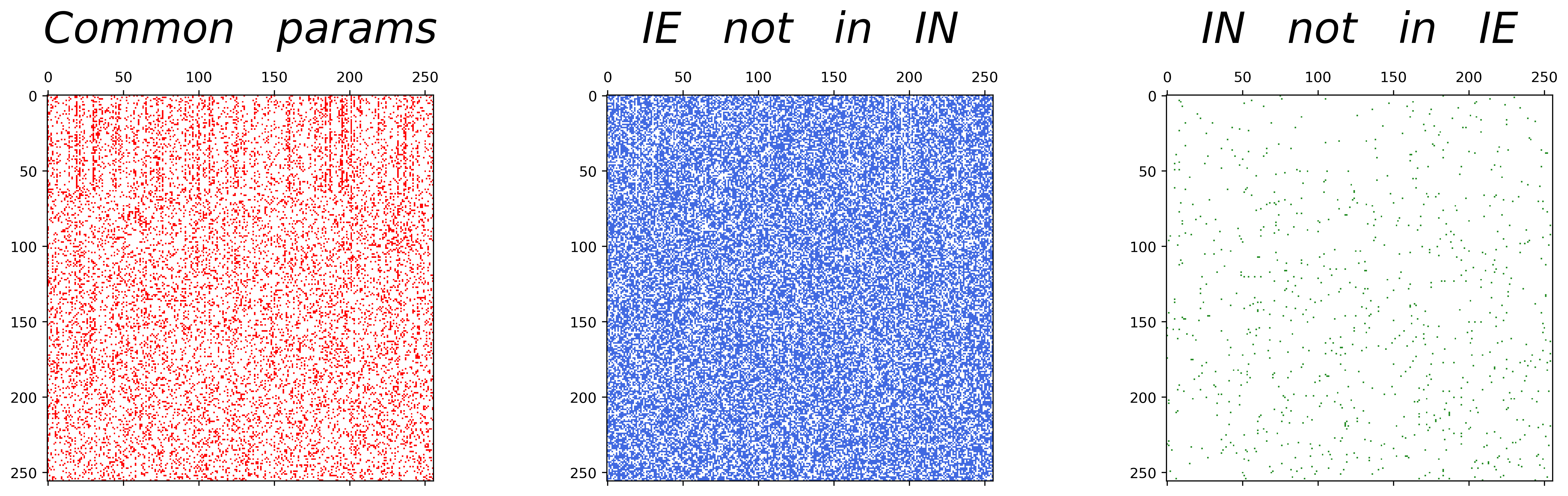}
            \caption{Key matrices}
        \end{subfigure}
        \hfill
        \begin{subfigure}[h]{0.48\linewidth}
            \centering
            \includegraphics[width=\linewidth]{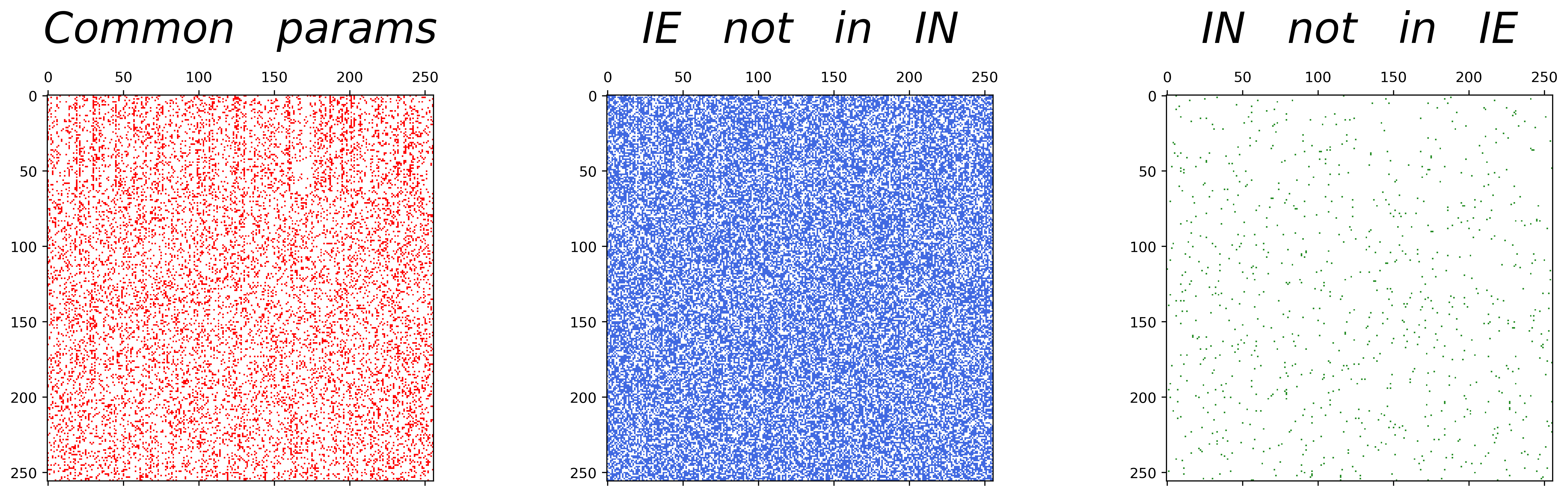}
            \caption{Query matrices}
        \end{subfigure}
        \hfill
        \begin{subfigure}[h]{0.48\linewidth}
            \centering
            \includegraphics[width=\linewidth]{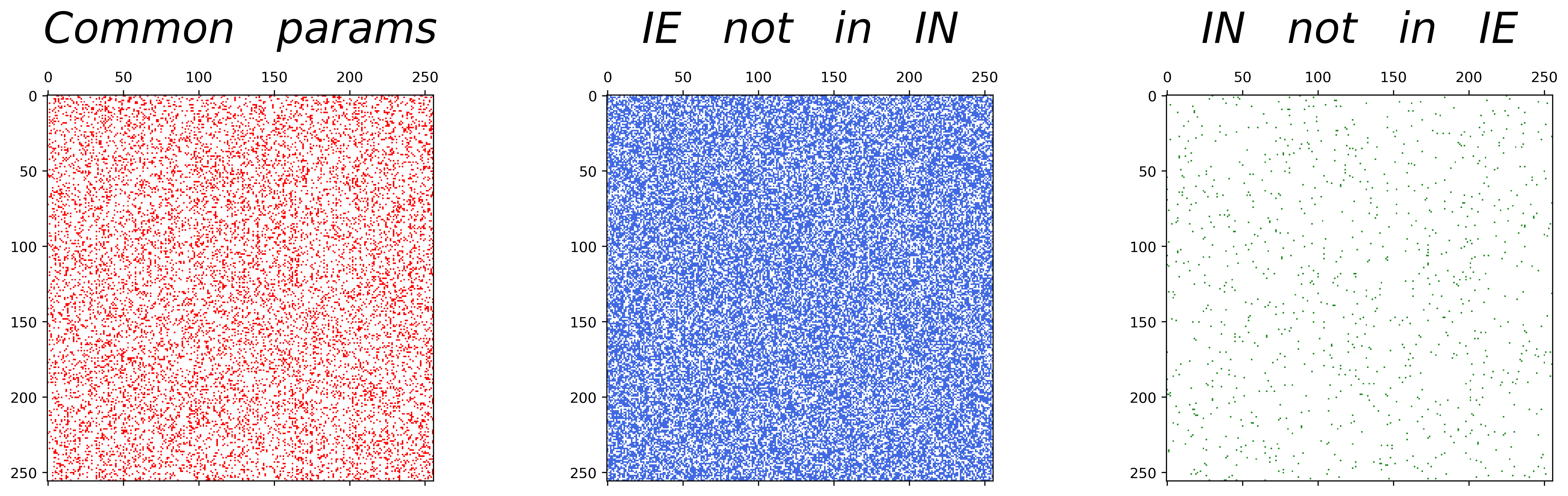}
            \caption{Value matrices}
        \end{subfigure}
    \caption{Layer 0 attention matrices}
\end{figure}

\begin{figure}[h!]
    \begin{subfigure}[h]{0.48\linewidth}
            \centering
            \includegraphics[width=\linewidth]{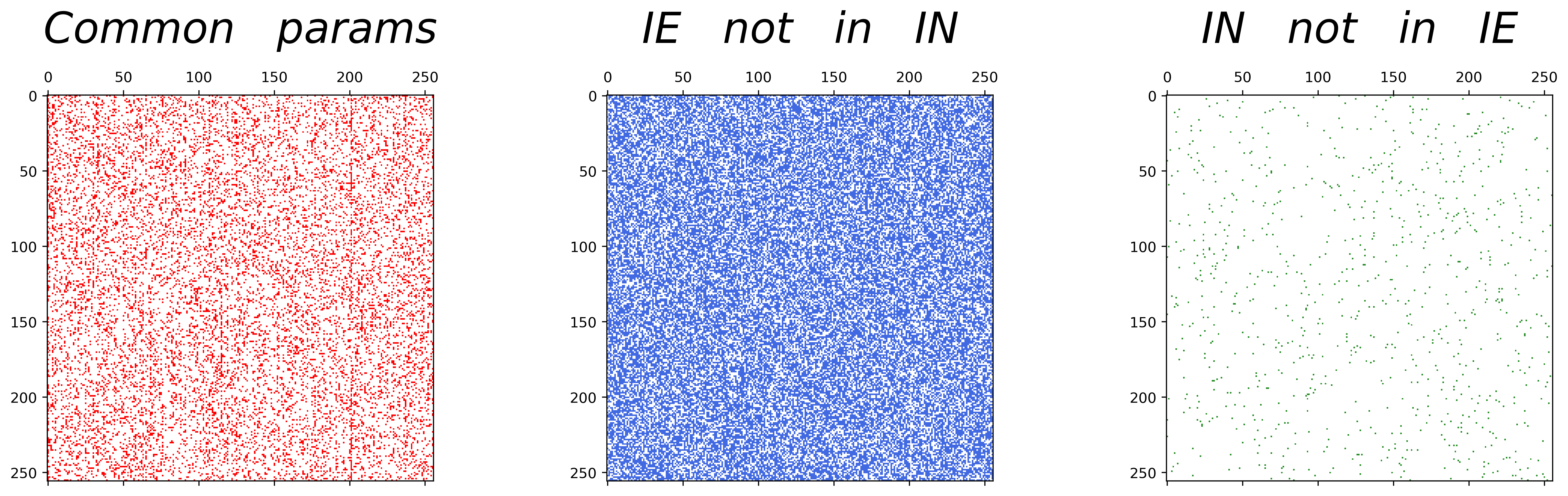}
            \caption{Key matrices}
        \end{subfigure}
        \hfill
        \begin{subfigure}[h]{0.48\linewidth}
            \centering
            \includegraphics[width=\linewidth]{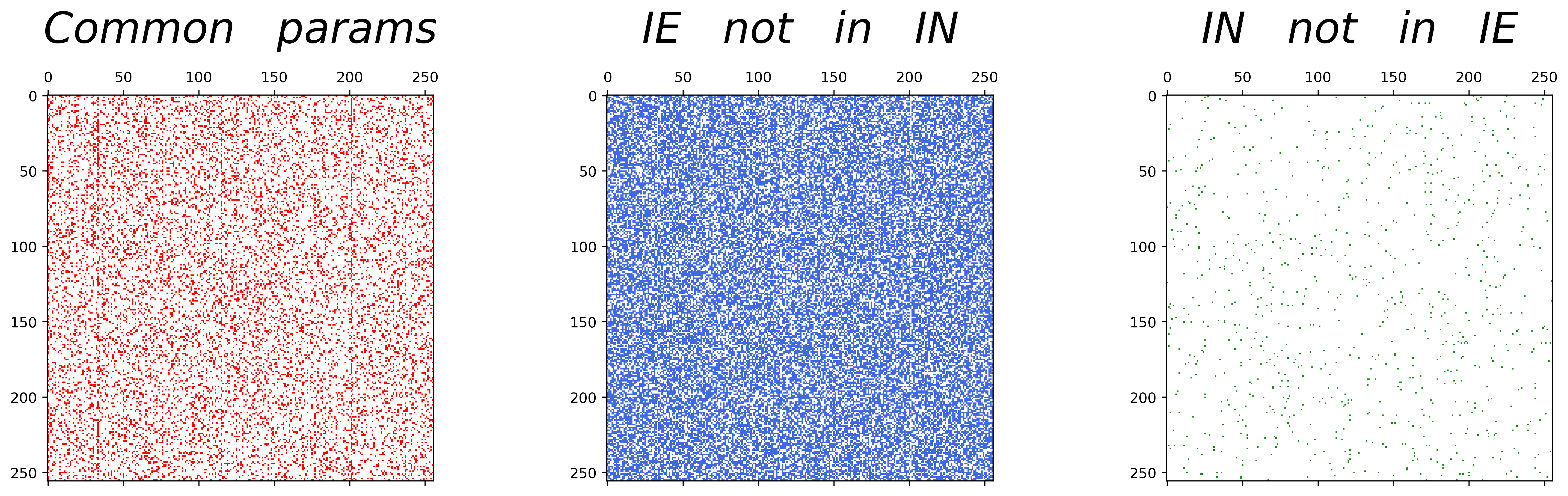}
            \caption{Query matrices}
        \end{subfigure}
        \hfill
        \begin{subfigure}[h]{0.48\linewidth}
            \centering
            \includegraphics[width=\linewidth]{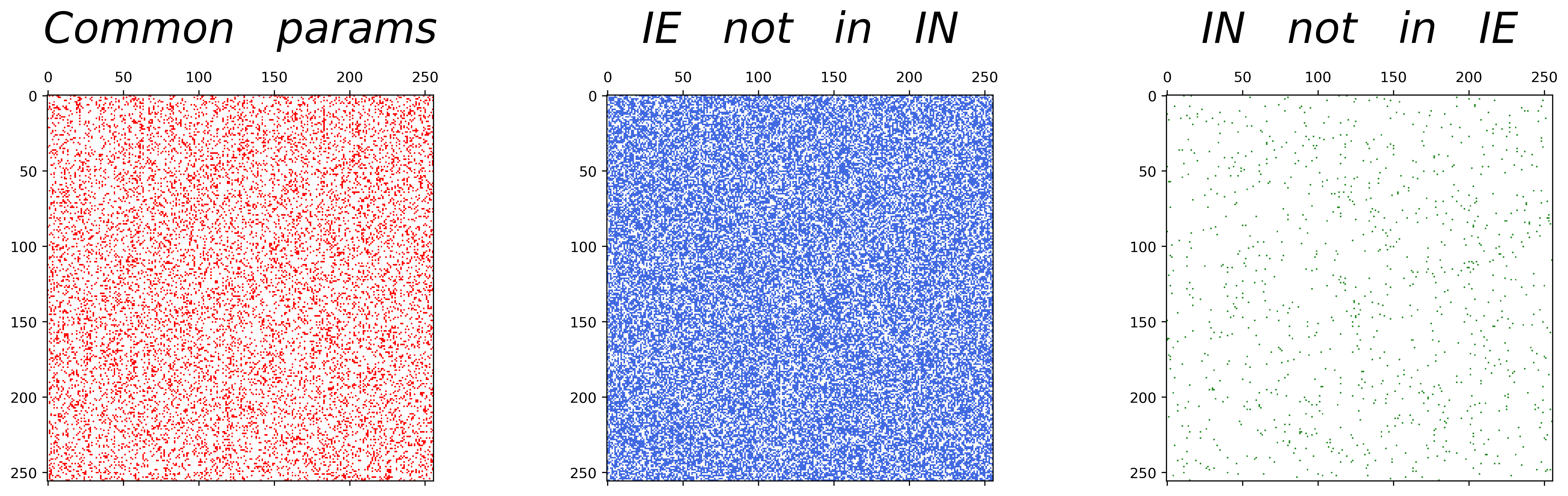}
            \caption{Value matrices}
        \end{subfigure}
    \caption{Layer 12 attention matrices}
\end{figure}
\newpage

\end{document}